\documentclass{article}

    \usepackage[final]{neurips_2025}

\usepackage[utf8]{inputenc} 
\usepackage[T1]{fontenc}    
\usepackage{url}            
\usepackage{booktabs}       
\usepackage{amsfonts}       
\usepackage{nicefrac}       
\usepackage{microtype}      
\usepackage{lipsum}

\usepackage{amsmath}
\usepackage{amssymb}
\usepackage{mathtools}
\usepackage{enumitem}
\usepackage{graphicx}  
\usepackage{amsthm}

\usepackage{multirow}       

\usepackage{url}
\usepackage[utf8]{inputenc} 
\usepackage[T1]{fontenc}    
\usepackage[breaklinks]{hyperref}      

\usepackage{booktabs}       
\usepackage{amsfonts}       
\usepackage{amsmath}
\usepackage[mathscr]{euscript}
\usepackage{amssymb}
\usepackage{mathtools}
\usepackage{amsthm}
\usepackage{nicefrac}       
\usepackage{microtype}      

\usepackage{graphicx}

\usepackage{caption}
\usepackage{multirow,makecell}

\usepackage{xspace}
\usepackage{subcaption}

\usepackage{algorithm}
\usepackage{algorithmic}
\usepackage{wrapfig}
\usepackage{listings}

\usepackage[capitalize,noabbrev]{cleveref}
\usepackage[textsize=tiny]{todonotes}

\theoremstyle{plain}

\theoremstyle{definition}

\theoremstyle{remark}

\usepackage{graphicx}
\usepackage{tikz}
\usepackage{adjustbox}

\definecolor{ForestGreen}{HTML}{228B22}

\title{
    \noindent
    \begin{adjustbox}{valign=c, margin=0pt 0pt}
        \begin{tikzpicture}
            \clip (0,0) circle (0.65cm);
            \node at (0,0) {\includegraphics[height=1.3cm]{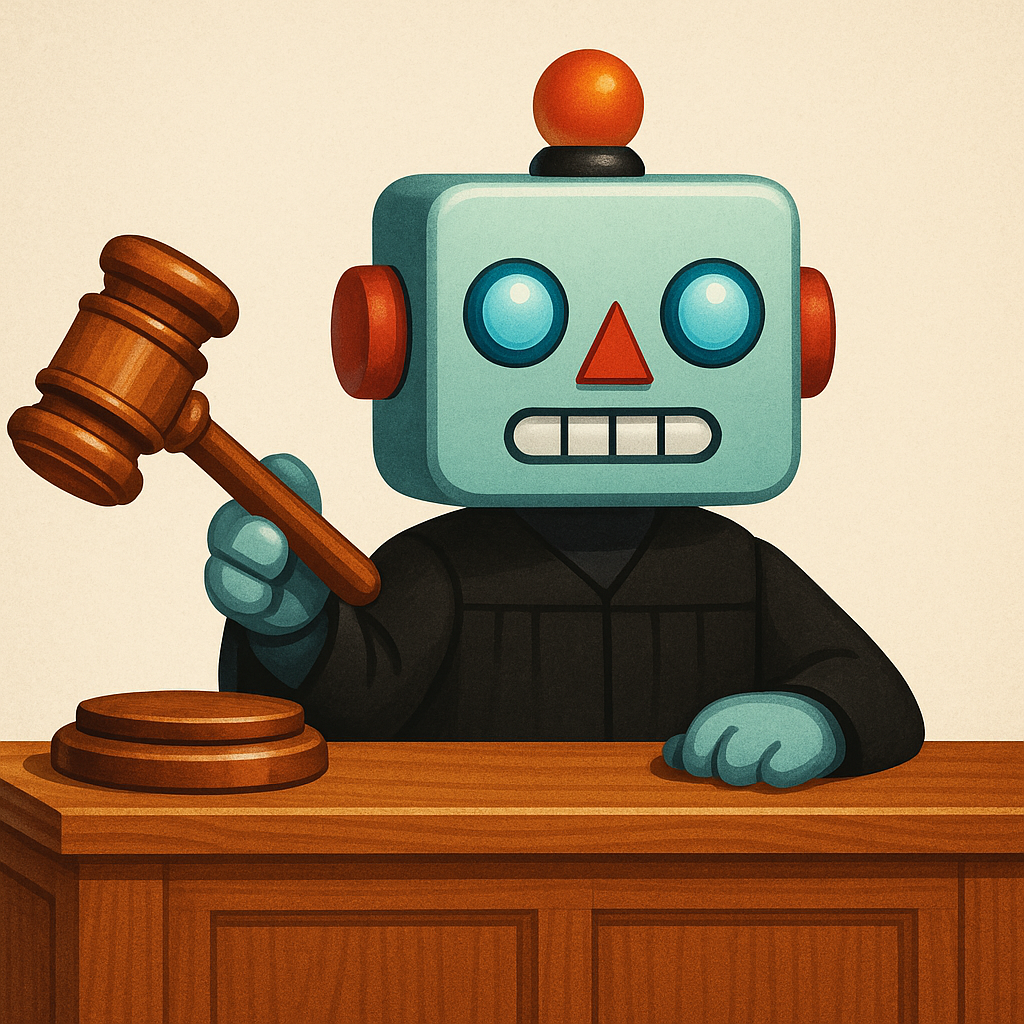}};
        \end{tikzpicture}
    \end{adjustbox}
    \hspace{-1.3cm}
    \begin{minipage}[c]{0.7\textwidth}
        \centering
        \textbf{AutoJudge: Judge Decoding} \\
        \textbf{Without Manual Annotation}
    \end{minipage}
    \hspace{-0.65cm}
}

\author{%
  Roman Garipov\,${}^{*\,\dagger}$ \\
  HSE University, Yandex \\
  \And
  Fedor Velikonivtsev\,${}^*$ \\
  HSE University, Yandex \\
  \And
  Ivan Ermakov\, \\
  HSE University, Yandex  \\
  \And
  Ruslan Svirschevski \\
  Yandex\\
  \And
  Vage Egiazarian ${}^{{\ddag}}$  \\
  IST Austria \\
  \And
  Max Ryabinin\\
  Together AI\\
}

\begin{document}

\maketitle

\begin{abstract}
We introduce AutoJudge\footnote[1]{Our code is available at \href{https://github.com/garipovroma/autojudge}{\texttt{github.com/garipovroma/autojudge}}.}, a method that accelerates large language model (LLM) inference with task-specific lossy speculative decoding. 
Instead of matching the original model output distribution token-by-token, we identify the generated tokens that affect the downstream quality of the response, relaxing the distribution match guarantee so that the ``unimportant'' tokens can be generated faster.
Our approach relies on a semi‑greedy search algorithm to test which of the mismatches between target and draft models should be corrected to preserve quality and which ones may be skipped.
We then train a lightweight classifier based on existing LLM embeddings to predict, at inference time, which mismatching tokens can be safely accepted without compromising the final answer quality.
We evaluate AutoJudge with multiple draft/target model pairs on mathematical reasoning and programming benchmarks, achieving significant speedups at the cost of a minor accuracy reduction. 
Notably, on GSM8K with the Llama 3.1 70B target model, our approach achieves up to ${\approx}2{\times}$ speedup \textit{over speculative decoding} at the cost of a ${\le} 1\%$ drop in accuracy.
When applied to the LiveCodeBench benchmark, AutoJudge automatically detects programming-specific important tokens, accepting ${\ge}25$ tokens per speculation cycle at a$~ 2\%$ drop in Pass@1. Our approach requires no human annotation and is easy to integrate with modern LLM inference frameworks.

\end{abstract}

\def\thefootnote{$*$}
\footnotetext{
    Equal contribution. ${}^\dagger$ 
    \hspace{-3px}Corresponding author: \texttt{devilgar@gmail.com}.
    \\
    ${}$\hspace{12px}${}^{{\ddag}}$ \hspace{-2px}Work done during employment at Yandex.
} 



\vspace{-10px}\section{Introduction}\label{sect:intro}\vspace{-5px}

Recent advances in LLM capabilities, including chain-of-thought reasoning~\citep{cot_wei_2022,zero_shot_cot_Kojima2022LargeLM,challenging_bigbench_solved_with_cot_Suzgun2022ChallengingBT}, writing complex software~\citep{Rozire2023CodeLO,Li2023StarCoderMT,swe_survey_2024}, or interacting with external tools~\citep{Schick2023ToolformerLM,Qin2023ToolLLMFL}, increasingly rely on inference-time computation~\citep{scaling_test_time_snell2024scaling,beeching2024scalingtesttimecompute}.
This progress is further accelerated with the release of reasoning-capable models, both proprietary~\citep{openai_o1,AnthropicClaude3.7Sonnet,googledeepmind2025gemini25thinking} and open-access~\citep{deepseek_r1,meta2025llama4,qwq32b}, that were explicitly trained to perform these kinds of inference-time computation. However, as LLMs tackle harder problems, they also tend to generate longer sequences~\citep{muennighoff2025s1} with tens of thousands of tokens~\citep{yeotong2025longcot}, taking up tens of minutes (and hundreds of dollars in costs) per task~\citep{openai_arc_prize_o3}.

A popular way to speed up LLM inference is through speculative decoding~\citep{leviathan2023fast,speculative_sampling_deepmind}, which uses a small ``draft'' model to propose the likely next tokens, then verifies these tokens with the main model in parallel. 
This method, along with its successors~\citep{specinfer,cai2024medusa,li2024eagle}, can speed up LLM inference while guaranteeing that the generated outputs match those of the original model (for greedy inference) or follow the same distribution. To achieve this, speculative decoding algorithms check if the draft tokens match the original model predictions. If there is a mismatch, they discard the incorrect token and all subsequent ones.

Speculative decoding can accelerate reasoning and other test-time computations, but it can be overly strict in how it discards tokens~\citep{bachmann2025judgedecoding,pan2025specreason,vivien2024optimal_lossy}. 
Intuitively, if a model generates a reasoning chain, not all mismatching tokens are equally important: errors in derivation should be fixed, while minor word choices should not.
Judge Decoding~\citep{bachmann2025judgedecoding} takes advantage of this by labeling which tokens are important for reasoning (and which are not) and allowing speculative decoding to accept more tokens by skipping the unimportant ones. However, their approach relies on human annotators to determine which tokens are important for reasoning. This complicates adoption and can be prone to human errors, particularly if the task requires expert knowledge (e.g., complex mathematical proofs or software engineering)

In this work, we look for ways to streamline this process. Instead of relying on human annotators, we propose \textbf{AutoJudge}, a search-based algorithm that detects which tokens are important for the task at hand based on how they affect the final answer. The algorithm is based on the idea that a token is deemed ``important'' not by itself, but in combination with other generated tokens. 
Thus, we propose a procedure that selects a small subset of important mismatching tokens that affect the final answer. 
Using this procedure, we can automatically mine a dataset to train an important token classifier that can then be used to accelerate speculative decoding.

The proposed search algorithm finds a small set of task-specific contextual ``important tokens'' --- cases where target and draft models disagree on the next token in a way that affects the final response quality. 
We then train a classifier to detect these important tokens and use it to improve traditional speculative decoding by relaxing its verification procedure.

Our experiments with Llama 3.x models demonstrate that the proposed approach can indeed identify important tokens and save time on speculation, accepting on average over 40 tokens per target model forward pass (approx. $2{\times}$ that of speculative decoding), at the cost of a ${\le}1{\%}$ drop in accuracy on GSM8K~\citep{cobbe2021gsm8k} and even more with a minor accuracy drawdown.
We obtain similar results with the Qwen2.5\citep{qwen2.5} family of models, observing comparable number of accepted tokens and accuracy trade-offs. 
When applied to programming tasks on LiveCodeBench~\citep{jain2024livecodebenchholisticcontaminationfree}, our approach is able to determine different task-specific important tokens, showing similar performance gains. The proposed framework is simple and general, using a classifier only when the original algorithm would reject a token, making it compatible with arbitrary speculative decoding algorithms. 
The main contributions of our work can be summarized as follows:
\begin{itemize}[leftmargin=*]
    \item We formulate AutoJudge, an algorithm for detecting which of the tokens generated in speculative decoding affect the downstream accuracy for a given task. Our algorithm requires no human annotation and can be applied to most popular LLM tasks.
    \item We verify the efficacy of AutoJudge on mathematical reasoning and programming benchmarks for several speculative decoding setups (model pairs). Our evaluations demonstrate favorable tradeoffs between the accuracy and the inference speedup, generating 20--45 tokens per speculative decoding cycle at the cost of a slight accuracy drawdown.
    \item We integrate AutoJudge with the vLLM framework~\citep{kwon2023efficient} and report the inference speed on A100 GPUs for both 8B and 70B target models, with up to $2{\times}$ speedup over speculative decoding at a ${\le}1\%$ quality decrease, and on H100 GPUs with a 405B target model.
\end{itemize}

\begin{figure}[t]
    \centering
    \includegraphics[width=1\linewidth]{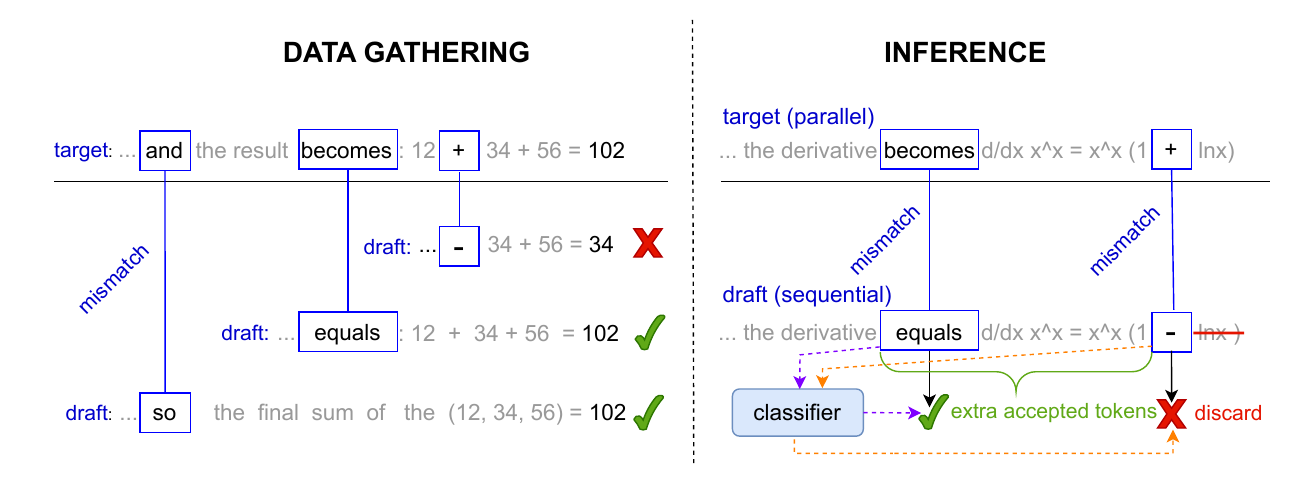}
    \vspace{-16px}
    \caption{Intuitive scheme of the proposed approach: \textbf{(left)} data collection: detecting mismatching tokens that affect final response quality; these tokens are then used to train a classifier \textbf{(right)} using the trained classifier to generate more tokens per cycle with speculative decoding.}
    \label{fig:layouts}\vspace{-10px}
\end{figure}
\vspace{-5px}
\section{Background}\label{sect:background}
\vspace{-5px}

\paragraph{Speculative Decoding.} Our work builds on top of speculative decoding~\citep{speculative_first,leviathan2023fast,speculative_sampling_deepmind}, a family of inference algorithms that accelerate token generation by improving hardware utilization. Speculative Decoding uses an auxiliary ``draft'' model to generate $K{>}1$ possible future tokens, then runs the main ``target'' model \textit{in parallel} to verify\footnote{For greedy decoding, it checks that the drafted tokens are the same as the target model's own next token predictions. For sampling, it uses a procedure that matches the sampling probabilities~\citep{leviathan2023fast}.} the generated tokens. The drafted tokens that agree with the target model predictions are accepted by the algorithm. In turn, the first mismatching token and all subsequent ones are rejected. This way, the method guarantees that all generated tokens follow the same distribution as sampling from the target model.
Subsequent works improve on this idea by generating draft trees instead of single sequences~\citep{specinfer,onlinespecdec,chen2024sequoiascalablerobusthardwareaware,specexec}, training specialized ``heads'' to draft next tokens based on the model's hidden states~\citep{cai2024medusa,ankner2024hydrasequentiallydependentdraftheads,li2024eagle,eagle2}, and more~\citep{spec_lookahead,speculative_staged,spectr,spec_retrieval}.\nocite{li2025eagle3scalinginferenceacceleration}

\vspace{-5px}\paragraph{Lossy Speculative Decoding.} The core guarantee of Speculative Decoding is that all generated tokens follow the probability distribution of the original model. However, there are practical scenarios where this guarantee can be sacrificed in favor of faster inference, which is known as lossy speculative decoding algorithms~\citep{vivien2024optimal_lossy,narasimhan2025fastercascades,biglittledecoder}.
Our work extends one such method: Judge Decoding~\citep{bachmann2025judgedecoding}\nocite{pan2025specreason}.
The core idea of Judge Decoding is that speculative decoding should only reject the mismatching token if accepting it would harm the response quality.
For instance, in mathematical reasoning, errors in the equations or logical fallacies are important for the final quality, while minor style changes are not. When writing code, algorithmic errors are important, while minor variable renames can be skipped in favor of faster inference.

The main challenge of Judge Decoding is determining which of the generated tokens can be skipped this way.
\cite{bachmann2025judgedecoding}~address this problem by manually labeling a training dataset for the classifier. Judge Decoding requires human annotators to find the ``mistake'' --- the first mismatching token that led the draft model to diverge from the original answer.
The resulting dataset of high-quality training examples is then used to train a linear classifier that detects such ``mistakes'' during inference.
Authors demonstrate that the collected dataset can, in principle, be reused across tasks and models. However, using the data from one task for inference on a different task results in substantial performance drawdown. Intuitively, different tasks (such as creative writing, math, or programming) have different criteria for which parts of the generated response matter most. Hence, it is best to train the important token classifier \textit{for the exact task at hand}. However, doing so with Judge Decoding would require relabeling the data by human annotators, which can be costly and time-consuming in specialized domains such as medicine or law. To alleviate this problem, we develop an automated search procedure for determining important tokens without external human (or LLM) annotators.
\vspace{-7px}
\section{Method Overview}\label{sect:method}
\vspace{-7px}

Our approach consists of three important stages. First, we detect which of the mismatching tokens affect the model quality using a semi-greedy search algorithm that we describe in Section~\ref{sect:method_mining}. We then use the gathered data to train a lightweight classifier that can detect important tokens at inference time (Section~\ref{sect:method_train_head}). Finally, we use the trained classifier to augment a speculative decoding algorithm as described in Section~\ref{sect:method_inference}, so that it can generate more tokens per speculation-verification cycle.

\vspace{-5px}
\subsection{Mining Important Tokens}\label{sect:method_mining}
\vspace{-5px}

In this section, we describe an algorithm to identify which draft tokens that mismatch with the target ones influence the final output quality. To achieve this, we systematically alter the generation output, swapping between draft and main model tokens and test how this affects the downstream task output, such as the final answer to a math problem or test outputs for a programming task.
If replacing a target model token with its draft version does not change the final answer, we deem this token swap ``unimportant'' and allow it to be generated with the faster draft model. In turn, if swapping out the token changes the final answer, it is deemed ``important'' and should be generated by the main model.

\newcommand{\comment}[1]{\color{violet}\(\triangleright\) #1\color{black}}  
\newcommand{\formulabox}[2][150]{\makebox[#1pt][l]{#2}}  
\newcommand{\commentmulti}[2][200]{\parbox[t]{#1pt}{\vspace{-7pt}\raggedright{\comment{#2}}}}  

\setlength{\textfloatsep}{5pt}
\begin{algorithm}[t]
\caption{\textsc{Search for Important Tokens}}\label{alg:important_tokens_mining}
\begin{algorithmic}[1]
\STATE \textbf{Input:} $x$: prompt, $\theta_\text{draft}$: draft model, $\theta_\text{target}$: target model
\STATE \textbf{Output:} a sequence of $\mathcal{M}$ mismatches, labeled as important or unimportant
\STATE $\mathcal{M} \gets \emptyset$\hspace{95px}\comment{A set of tuples (position, target token, draft token, important)}
\STATE $y \gets \textsc{generate}(x, \theta_\text{target})$
\STATE $\alpha \gets \textsc{extractAnswer}(y)$
\STATE $\widetilde{y}\gets \textsc{forward}(x \oplus y, \theta_\text{draft})\texttt{.argmax(-1)[len(x)-1:-1]}$
\STATE \formulabox{$\mathcal{I} \gets \{i \mid y_i \ne \widetilde{y}_i \}$}\hspace{23px}\commentmulti{Indices where draft and target tokens mismatch}
\WHILE{$\mathcal{I} \ne \emptyset$}
    \STATE \formulabox{$t \gets \min(\mathcal{I})$} \hspace{8px} \commentmulti{The earliest position where mismatch happened}
    \STATE \formulabox{$\hat y = y_{1:t} \oplus \widetilde y_{t} \oplus \textsc{generate} (x \oplus y_{1:t} \oplus \widetilde y_{t}, \theta_{target})$ \commentmulti{Replace $\widetilde y_t$ and continue with $\theta_{target}$}} 
    \STATE $\hat{\alpha} \gets \textsc{extractAnswer}(\hat{y})$
    \IF{$\alpha \equiv \hat{\alpha}$}
        \STATE \formulabox{$\mathcal{M} \gets \mathcal{M} \cup \{(t, y_t, \widetilde{y}_t, \texttt{False})\}$} \commentmulti{Equivalent answer, token $y_t$ is \underline{not important}}
        \STATE $y \gets \hat{y}$ \hspace{123.5px}\commentmulti{Continue search from the new response}
        \STATE $\widetilde{y}\gets \textsc{forward}(x \oplus y, \theta_\text{draft})\texttt{.argmax(-1)[len(x)-1:-1]}$
    \ELSE
        \STATE \formulabox{$\mathcal{M} \gets \mathcal{M} \cup \{(t, y_t, \widetilde{y}_t, \texttt{True})\}$} \comment{Different answer, token $y_t$ is \underline{important}, keep it}
    \ENDIF
    \STATE $\mathcal{I} \gets \{i | y_i \neq \widetilde y_i \; \cap \; i > t \}$ \hspace{53px} \commentmulti{Continue with the remaining mismatches after $t$}
\ENDWHILE
\RETURN $\mathcal{M}$
\end{algorithmic}
\end{algorithm}

In more formal terms, consider the task defined as a prompt $x$ with and two models: the larger $\theta_{target}$ and the smaller $\theta_{draft}$. Both models can generate a response $y=(y_1, \dots, y_T)=\textsc{generate}(x, \theta_{draft})$ with up to $T{\le}T_{max}$ total tokens.
For simplicity, we first assume that the \textsc{generate} procedure is deterministic (e.g., greedy) and generalize to sampling in Appendix~\ref{app:algo_details}.

Without the loss of generality, we also assume that there is a problem-specific way to extract the final answer from the model's response, $a=\textsc{extractAnswer}(y)$. In mathematical reasoning tasks such as GSM8K~\citep{cobbe2021gsm8k}, the final answer is literally whatever the model puts after \texttt{"the final answer (is)"}. In programming tasks, the ``answer'' would be the output from the testing system given the generated code --- either a report about passing and failing tests or a testing error (e.g., an Out Of Memory or Syntax Error). Finally, we say that two answers are equivalent $a_{ref}{\equiv}a_{alt}$ if they are the same from the downstream task perspective. Note that this does not require them to be exactly equal as strings: in math problems, $\texttt{1.5}{\equiv}\texttt{3/2}$, whereas in programming tasks, two programs can be equivalent despite having different variable names. If the task at hand does not have a formalized evaluation procedure, e.g., general conversation agents, we can define $\textsc{extractAnswer}(y) = y$ and detect if two answers are equivalent using an LLM or human judges.

Following this notation, let $y_{target} = \textsc{generate}(x, \theta_{target})$ be the main model outputs. A token $y_t \in y_{target}$ is \textbf{un}important if swapping that token for the draft model's output results in an equivalent answer. Likewise, if replacing $y_i$ (and continuing target generation from there) results in a different answer, then the original token was ``important'' and the token should be generated with $\theta_{target}$.

Note that even if $\theta_{draft}$ is significantly smaller than $\theta_{target}$, most of the individual tokens will match between the two. As such, we are only interested in the mismatches --- the cases where draft and target models produce different tokens \textit{given the same prefix}:

\vspace{-15px}
\begin{equation*}    
\mathcal{I}(x) = \{ t \in [1, T) : \underset{y_{next}}{\arg\max} \; P(y_{next} | x, y_{1:t}, \theta_{target}) \neq \underset{y_{next}}{\arg\max} \; P(y_{next} | x, y_{1:t}, \theta_{draft})\},
\end{equation*}
\vspace{-10px}

where $y_{1:t}=y_1, \dots, y_{t-1}$ denotes taking a prefix of $y$ up to, but excluding index $t$.
In practice, we can find these tokens quickly by re-encoding the target model response with the draft model: $\textsc{forward}(x \oplus y, \theta_{draft}).\texttt{argmax(dim=-1)[M{-}1:M{+}T{-}1]}$
, where $x \oplus y$ denotes concatenation,
$\textsc{forward}(\cdot, \cdot)$ is a parallel transformer forward pass that outputs next token logits,
and the $\texttt{logits}.\texttt{argmax(dim=-1)[M{-}1:M{+}T{-}1]}$ takes the most likely next tokens for every position, excluding the prompt and accounting for the shift from next token prediction.

When deciding if a mismatching token is important for the final response, we need to account for the fact that changing one token will most likely lead to changes in subsequent tokens. A na\"ive way to account for that change is by continuing\footnote{For notation simplicity, we assume that the $\textsc{generate}(\cdot, \cdot)$ function can be called with a prefix of a response. In that case, we assume that the \textit{total} response length (and not just newly generated tokens) does not exceed $T_{max}$, so that the response cannot grow indefinitely with each subsequent replacement.} the response after replacing one token $\widetilde y_{t}$:

\vspace{-12px}
\begin{equation*}
\hat y = y_{1:t} \oplus \widetilde y_{t} \oplus \textsc{generate} (x \oplus y_{1:t} \oplus \widetilde y_{t}, \theta_{target})
\end{equation*}

However, this approach has a significant downside in that it assumes that all subsequent tokens will be generated by $\theta_{target}$, whereas in reality, some of them may be generated by $\theta_{draft}$ following the same algorithm. In preliminary experiments, we found that, with a capable enough $\theta_{target}$, even significant generation errors can be detected and self-corrected (similar to the ``Aha moment'' from~\cite{deepseek_r1,muennighoff2025s1}). However, if the model makes multiple mistakes, they eventually reach a critical mass, leading to an incorrect answer.

To address this, we reframe our task from detecting individual important tokens to finding combinations of tokens that jointly affect the final answer. This changes our problem to \textbf{finding the minimal set of mismatching tokens that need to be generated by $\theta_{target}$ while still producing an equivalent answer}\footnote{More precisely, find the fastest-to-generate sequence, accounting for the differences in response length.}. Since replacing a single mismatching token affects all subsequent token choices, the exact solution to this problem requires a tree search over possible token assignments. While this type of tree search is possible, it would take up significant runtime due to the large number of LLM forward passes required to try all mismatch combinations.

To simplify the procedure, we opt instead for a simpler, semi-greedy search that starts from the target model response and iteratively tries to replace mismatching target model tokens with their draft counterparts. If replacing a token affects the final answer, we consider this token important and keep the original (target model) version. If, however, replacing the token results in an equivalent answer, we deem this token unimportant, replace it with the draft model version \textit{and continue the search from the new sequence}, with a different suffix and possibly a different $\mathcal{I}$. That way, we guarantee that the search algorithm is aligned with what happens during inference: the important tokens are generated with the target model and the unimportant ones are kept from the draft model. We summarize the resulting search procedure in Algorithm~\ref{alg:important_tokens_mining} and discuss some of its implications in Appendix~\ref{app:algo_details}.

\subsection{Classifier Training}\label{sect:method_train_head}

\begin{figure}[t]
    \vspace{-5px}
    \textcolor{gray}{[GSM8K] \texttt{Arnel\! had\! ten\! boxes of\! pencils ...\! how\! many\! pencils\! are\! in\! each\! box?}}
    
    \textcolor{black}{\texttt{Arnel kept ten pencils \textcolor{ForestGreen}{\textbf{and}} shared the remaining pencils \textcolor{ForestGreen}{\textbf{with}} his 5 friends.}}
    
    \textcolor{gray}{\;\texttt{\hspace{117px}\textbf{\textcolor{ForestGreen}{[.]}} He shared the ... \textcolor{ForestGreen}{\checkmark} \hspace{18px} \textbf{\textcolor{ForestGreen}{[equally]}} with ...} \textcolor{ForestGreen}{\checkmark}}
    
    \textcolor{black}{\texttt{This means that the total number of pencils \textcolor{ForestGreen}{\textbf{he}} shared is 10 \textcolor{red}{\textbf{*}} x - 10. ...}}
  
    \textcolor{gray}{\;\texttt{\hspace{215px}\textbf{\textcolor{ForestGreen}{[Arnel]}} ... \textcolor{ForestGreen}{\checkmark} \hspace{4px} \textbf{\textcolor{red}{[-]}}\! x\! -\! 10 ...} \textcolor{red}{$\bigtimes$}}

    \vspace{-5pt}
    \noindent\rule{\linewidth}{0.5pt} 
    
    \vspace{-1px}
    
    \begin{minipage}{0.499\textwidth}
    \vspace{10px}
    \textcolor{gray}{[GSM8K]\! \texttt{Adlai\! has\! 2\! dogs\! and\! 1\! chicken.\,\,\, How\! many animal\! legs are there\! in\! all?}}

    \texttt{To\! find\! the\! total\! number\! of\! animal\! legs, we need to calculate the \textcolor{ForestGreen}{legs}$\,_{\textcolor{ForestGreen}{\texttt{[total]}}}$ of each animal and then add them up.}

    \vspace{7px}

    \texttt{- 2 dogs have \textcolor{red}{4}$\,\,{}_{\textcolor{red}{[2]}}$ legs each, so 2 dogs have 2 \textcolor{ForestGreen}{*}$\,{}_{\textcolor{ForestGreen}{\texttt{[times]}}}$ 4 = 8 legs.}
    
    \texttt{- 1 chicken has 2 legs.}

    \vspace{7px}

    \texttt{\textcolor{ForestGreen}{Now}$\,{}_{\textcolor{ForestGreen}{\texttt{[Adding]}}}$, let's add the legs \textcolor{ForestGreen}{together}$\,{}_{\textcolor{ForestGreen}{\texttt{[of]}}}$
    \!, we get 8\! (from the dogs) + 2\! (from the chicken) = 10 legs.
    }

    \vspace{7px}

    \texttt{The final answer is 10.}

    \vspace{1px}

    \end{minipage}
    \hfill\vline\hspace{3px}\hfill
    \begin{minipage}{0.48\textwidth}
    \vspace{10px}

    \textcolor{gray}{[LCB]\! \texttt{Given\! a\! string\! S\! of\! lowercase...\!\!\!\!
If\! there\! are\! adjacent\! occurren-\!\! ces\! of a and b in S, print Yes;\! ...\!}}
    
    \texttt{\`{}\`{}\`{}python}
    
    \texttt{\# \textcolor{ForestGreen}{-*-$_{[\texttt{YOUR}]}$} coding: utf-8 -*-}

    \texttt{\textcolor{ForestGreen}{def$_{[\texttt{\#}]}$ solve$_{[\texttt{check}]}$}(s):}
    
    \texttt{\quad for i in range(len(s) \textcolor{red}{-$_{[)]}$} 1):}

    \texttt{\quad\quad if s[i]\! ==\! 'a'\! and\! s[i+1]\! ==\! 'b':}

    \texttt{\quad\quad\quad return "Yes"}

    \texttt{\quad\quad if s[i]\! ==\! '\textcolor{red}{b$_{[a]}$\!}'\! and\! s[i+1]\!\! ==\!\! 'a':\!\!\!\!\!\!\!\!\!\!\!\!\!\!\!\!\!\!\!\!\!}

    \texttt{\quad\quad\quad return "Yes"}

    \texttt{\quad return "No"}

    \vspace{7px}
    
    \texttt{\textcolor{ForestGreen}{if$_{\texttt{[\#]}}$} \_\_name\_\_ == "\_\_main\_\_":}\textcolor{gray}{\texttt{...}}
    \end{minipage}
    \vspace{3px}
    \caption{Excerpts from GSM8K (top, left) and LiveCodeBench (right) labeled by Algorithm~\ref{alg:important_tokens_mining}. Important mismatching tokens that are in \textcolor{red}{red}, unimportant ones are in \textcolor{ForestGreen}{green}. Alternative tokens are shown in \texttt{[}brackets\texttt{]}. Black tokens are where $\theta_{draft}$ and $\theta_{target}$ gave the same prediction. The top example additionally shows $\theta_{target}$ continuations after mismatching tokens (\textcolor{ForestGreen}{$\checkmark$} if $\alpha \equiv \hat{\alpha}$, \textcolor{red}{$\bigtimes$} if not).}
    \label{fig:examples}
    \vspace{5px}
\end{figure}

Once we gather a dataset of task-specific important tokens with \cref{alg:important_tokens_mining}, we can train a classifier that would detect such tokens for use during inference. This classifier can, in principle, be any type of model, from a simple linear model or decision tree to a fine-tuned transformer layer. However, in our work, we default to training lightweight \textbf{linear models with existing LLM hidden states as features}, since those would introduce the least overhead during inference. There are several important design choices that can affect the effectiveness of such a classifier: we address each one separately.


\textbf{1. Which token representations to use:} the hidden states that predicted the mismatched token, or the next hidden states that encode the mismatched token itself? In our experiments, we found that using the latter representations results in substantially greater classifier accuracy (see Appendix~\ref{app:classifier_inputs}). However, obtaining these representations comes with a caveat.

Normally, when doing speculative decoding, one generates a draft ``window'' of $W$ tokens with $\theta_{draft}$, then verifies these tokens by processing them (in parallel) with $\theta_{target}$. This automatically computes the necessary hidden representations for all but for the very last token --- the next token predicted from the last hidden state in the window, which is not encoded. There are two ways to address this: either encoding the extra token alongside the window, or simply assuming that \underline{if} the very last token mismatches between $\theta_{draft}$ and $\theta_{target}$, it is automatically discarded without the classifier. However, in practice, we found that the overhead from either strategy is negligible and is outweighed by greater classifier accuracy that translates to more accepted tokens.

\textbf{2. Which token alternative to use?} Since the classifier works best with the representations from encoding the mismatching token, it is natural to ask which token should be encoded: the draft token, the mismatching target token, or both? When analyzing this, we found that using both token representations comes with a \textit{slight} increase in classifier accuracy (see Appendix~\ref{app:classifier_inputs}). However, obtaining these representations in practice would require running $\theta_{target}$ more than once during the verification stage, which would complicate inference and introduce performance overhead. For this reason, we opt to use only the draft token representations for the classifier, since those are already available in regular speculative decoding.

\vspace{-1px}\textbf{3. Which model provides feature representations?} During the verification stage, we have access to both the draft and the target model hidden states: we can use either or both of them as the input. In practice, we found that concatenated draft and target model representations give slightly better results than just those of the target model, and using draft model representations alone is substantially worse. Since both representations are already available during inference, we opt to use each of them.

\vspace{-1px}\textbf{Classifier model \& training.} In this work, we train a simple logistic regression to detect important tokens. While a more complex model could achieve greater accuracy, logistic regression is significantly easier to deploy, has less runtime \& memory overhead and needs less training data. Furthermore, it can be fused with the existing ``LM head'' layer of the draft and target LLMs, which would make its computation virtually free. To control overfitting, we perform a simple grid search over the $L_2$ regularization coefficient (``$C$'') with a logarithmic grid. We report additional details in Appendix~\ref{app:classifier_inputs}.

\vspace{-5pt}
\subsection{Inference}\label{sect:method_inference}
\vspace{-5pt}

The resulting classifier can be used with an arbitrary speculative decoding algorithm that has a verification stage. During said verification stage, the classifier is called when the original algorithm would reject a token. If the would-be-rejected token is deemed to be unimportant, i.e. not to affect the response quality, then we override the verification procedure and accept the token instead, proceeding to test subsequent tokens (if any) as per the original algorithm.

\vspace{-1px}\textbf{Generality.} In our initial experiments, we focus on traditional speculative decoding~\citep{leviathan2023fast,speculative_sampling_deepmind} for simplicity.
However, our algorithm is compatible with arbitrary speculative decoding algorithms, including tree-based~\citep{specinfer,specexec,chen2024sequoiascalablerobusthardwareaware} and single-model multi-head algorithms~\citep{cai2024medusa,li2024eagle,eagle2}. This also means that our approach can be integrated into existing inference frameworks such as vLLM~\citep{kwon2023efficient}, TensorRT-LLM~\citep{trt-llm} or TGI~\citep{tgi}.

\vspace{-1px}\textbf{Thresholds.} To balance computational efficiency and downstream performance, we select a decision threshold that achieves a high recall (${\ge}90{\%}$) in order to retain quality.
Since the classifier is accurate enough, this threshold can also achieve a reasonable rejection rate, i.e., the rate of tokens correctly predicted to be unimportant. 
This allows us to retain downstream accuracy while skipping a large portion of unimportant tokens, thus enabling efficient speculative decoding. 
In Section~\ref{sect:experiments}, we evaluate various threshold values to show their effect on accuracy and acceptance rate.

\vspace{-1px}\textbf{Comparison with Judge Decoding.} As we discussed earlier, our approach can be seen as an extension of Judge Decoding that enables automatic dataset mining. 
As such, the dataset generation algorithm from Section~\ref{sect:method_mining} can be used in conjunction with the Judge Decoding training and inference protocol, which appears to be similar to ours up to possible minor details. 
In Appendix~\ref{app:manual_annotation}, we additionally compare against manual human annotation similar to Judge Decoding.
Unfortunately, the original source code and data from Judge Decoding are not available, making it difficult to compare directly.

\section{Experiments}\label{sect:experiments}
\vspace{-5px}

\begin{figure}[t]
    \vspace{-10px}
    \centering
    \includegraphics[width=0.49\linewidth]{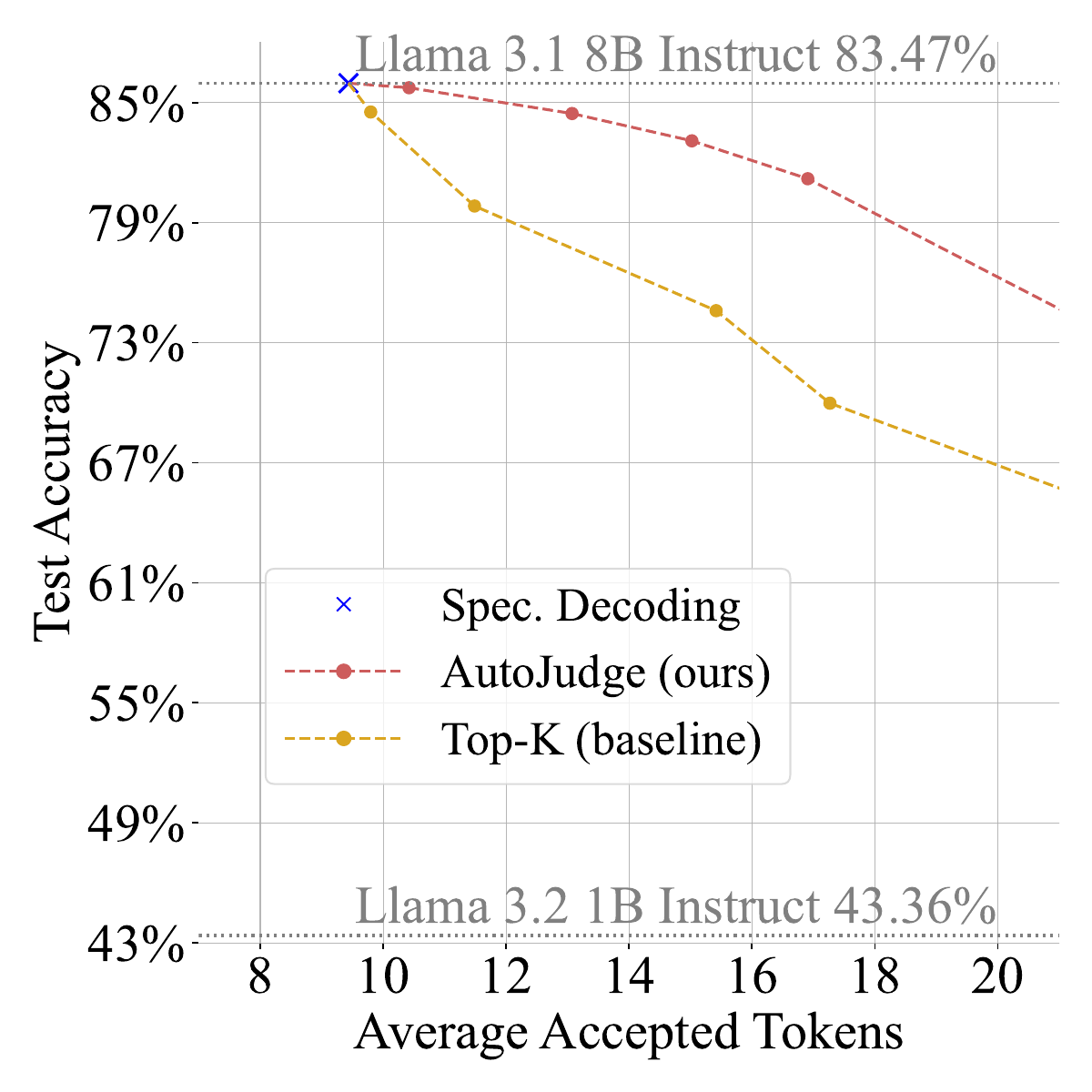} 
    \includegraphics[width=0.49\linewidth]{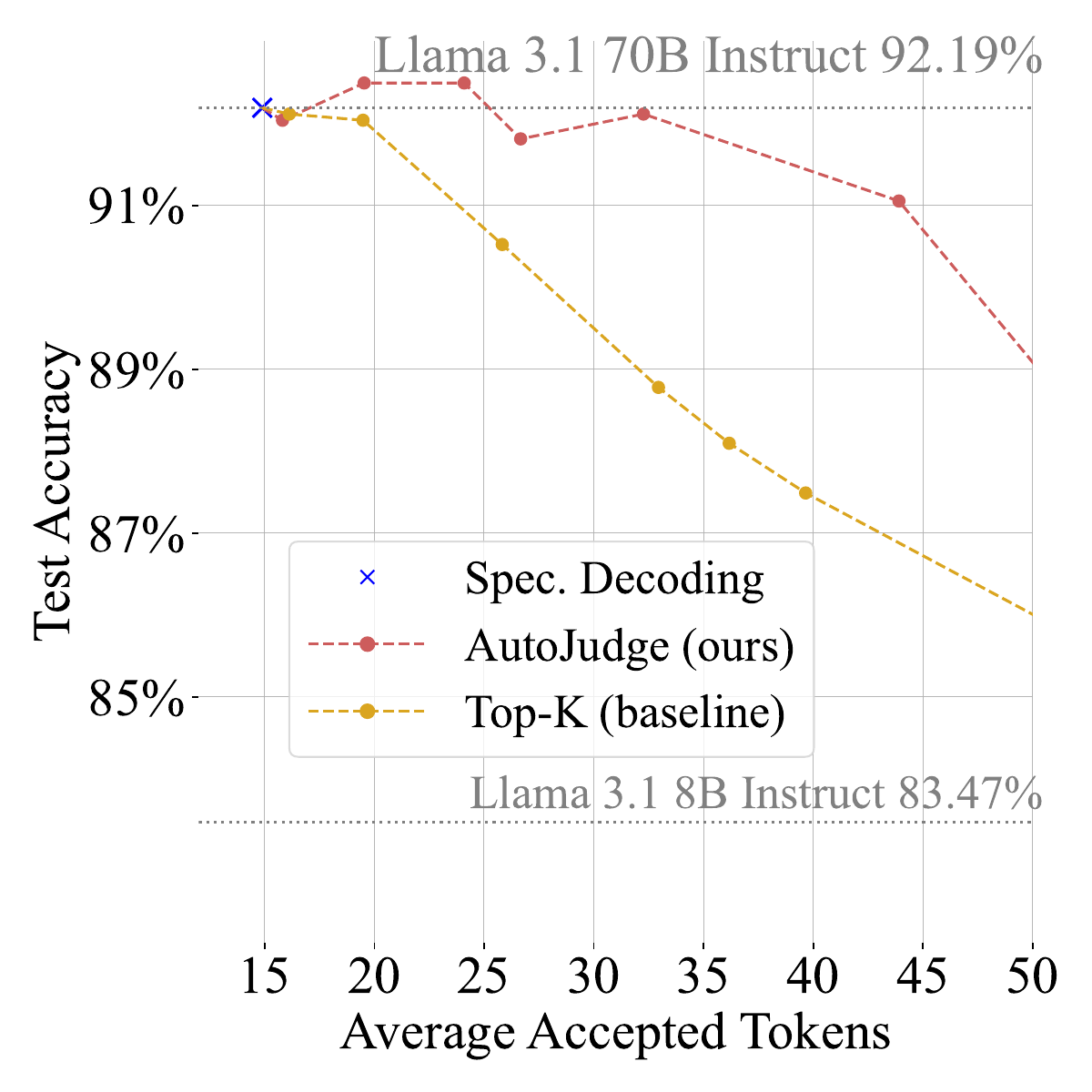}
    \caption{Accuracy and the number of accepted tokens on GSM8K for \textbf{(left)} 8-shot Llama-3.2 1B draft / Llama-3.1 8B target and \textbf{(right)} 0-shot Llama 3.1 8B draft / Llama 3.1 70B target (all Instruct)}
    \label{fig:main_plots_gsm8k}
    \vspace{5px}
\end{figure}

We organize our evaluations as follows: in Section~\ref{sect:exp_math_gsm8k} we evaluate AutoJudge on the GSM8K~\citep{cobbe2021gsm8k} mathematical reasoning benchmark. 
Next, in Section~\ref{sect:exp_coding_lcb}, we evaluate our approach on programming tasks from LiveCodeBench~\citep{jain2024livecodebenchholisticcontaminationfree}.
Finally, Section~\ref{sect:exp_vllm_inference} contains GPU inference speed benchmarks with our vLLM implementation.
We focus on two pairs of Llama 3.x models: 1) Llama-3.2-1B-Instruct draft / Llama-3.1-8B-Instruct target\footnote{The reason why the two models have different minor versions (i.e. 3.1 and 3.2) is that the 3.2 version does not have the larger 8B models, and the 3.1 version does not have the smaller 1B models.} and 2) Llama-3.1-8B-Instruct draft / Llama-3.1-70B-Instruct target. We also report results with Qwen2.5 models on the GSM8K benchmark to showcase the method’s transferability across model families.
Our main experiments run in \texttt{bfloat16} precision (see Appendix~\ref{app:precision}).
We run AutoJudge on top of the standard speculative decoding algorithm~\citep{leviathan2023fast} in main experiments and explore EAGLE-2 in Appendix~\ref{app:eagle}.

\vspace{-5px}
\subsection{Mathematical Reasoning with GSM8K}\label{sect:exp_math_gsm8k}
\vspace{-5px}

Our first set of experiments is based on the GSM8K dataset with grade school mathematical problems. 
This dataset has a natural split with ${\approx}7.47$K training samples and ${\approx}1.32$K test samples. 
Following the standard evaluation procedure, we use the training set to ``mine'' important tokens with Algorithm~\ref{alg:important_tokens_mining} and train the classifier, then run inference and evaluate on the test set with the recommended parameters~\citep{eval-harness} for zero-shot and 8-shot evaluation: greedy inference with a prompt that encourages chain-of-thought reasoning.
During training, we consider two responses equivalent ($a{\equiv}\hat a$ in \cref{alg:important_tokens_mining}) if the extracted final answers (numbers) are equal. For reference, we provide an example of important token assignments found by our algorithm in Figure~\ref{fig:examples}.

We train a classifier on the last hidden state embeddings from both draft and target models (concatenated) for encoded draft tokens. The training dataset contains ${\approx}130$K mismatches, about $20\%$ of which are deemed important. We train a logistic regression with the $L_2$ regularization coefficient tuned individually for each setup(for instance $10^{-4}$ for 8B/70B), found by grid search over a logarithmic grid between $10^{0},\dots, 10^{-7}$.
During inference, we integrate the trained classifier into the speculative decoding loop from~\cite{leviathan2023fast} during verification. 
Whenever the original algorithm would reject a token, we run the classifier to determine if changing that token affects the final response quality, and if not --- accept the token and continue verification for subsequent tokens (if any). 
Since the resulting algorithm can accept additional tokens, we use the increased draft window size of $W{=}64$ tokens for all evaluations.

\begin{figure}[t]
    \vspace{-10px}
    \centering
    \includegraphics[width=0.49\linewidth]{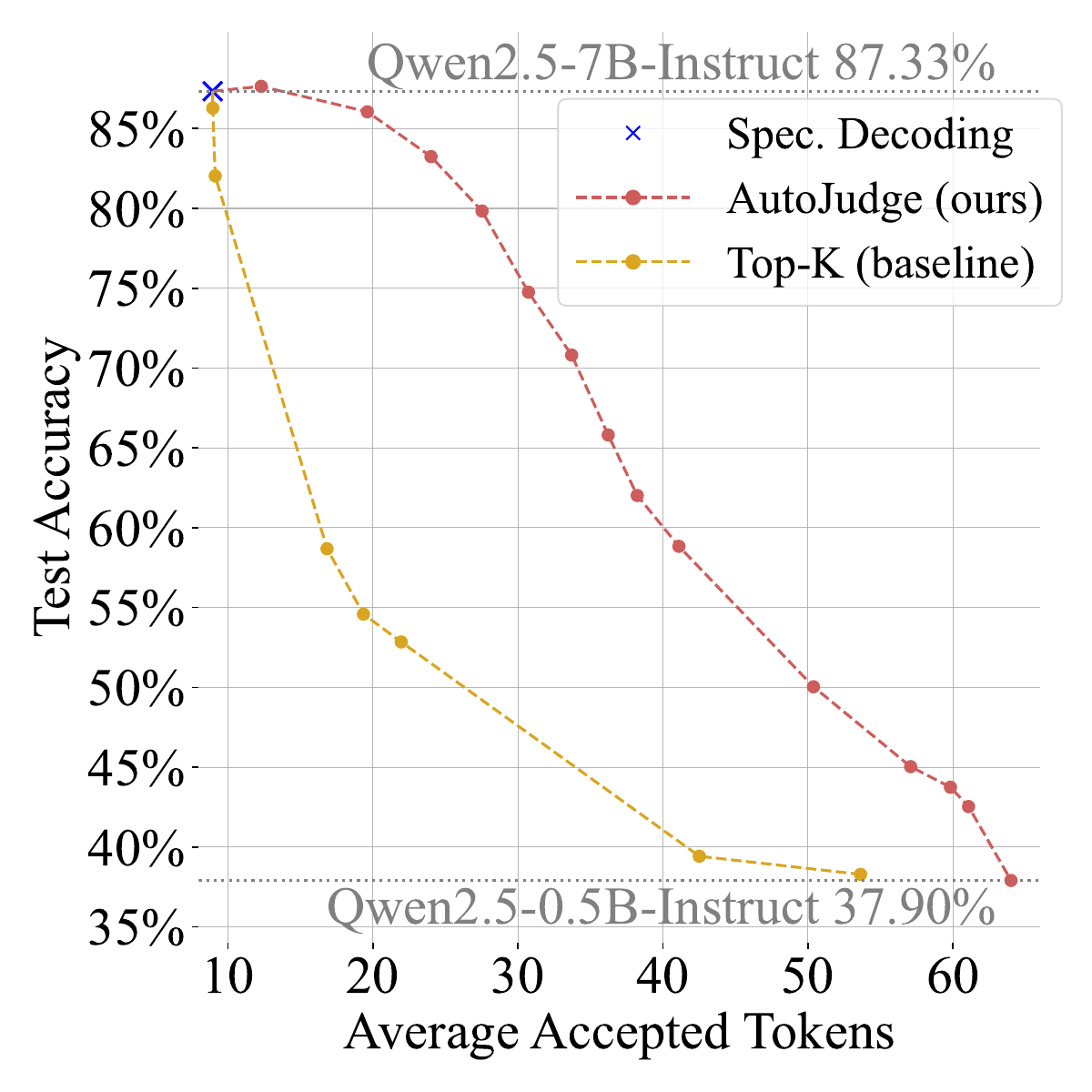}
    \includegraphics[width=0.49\linewidth]{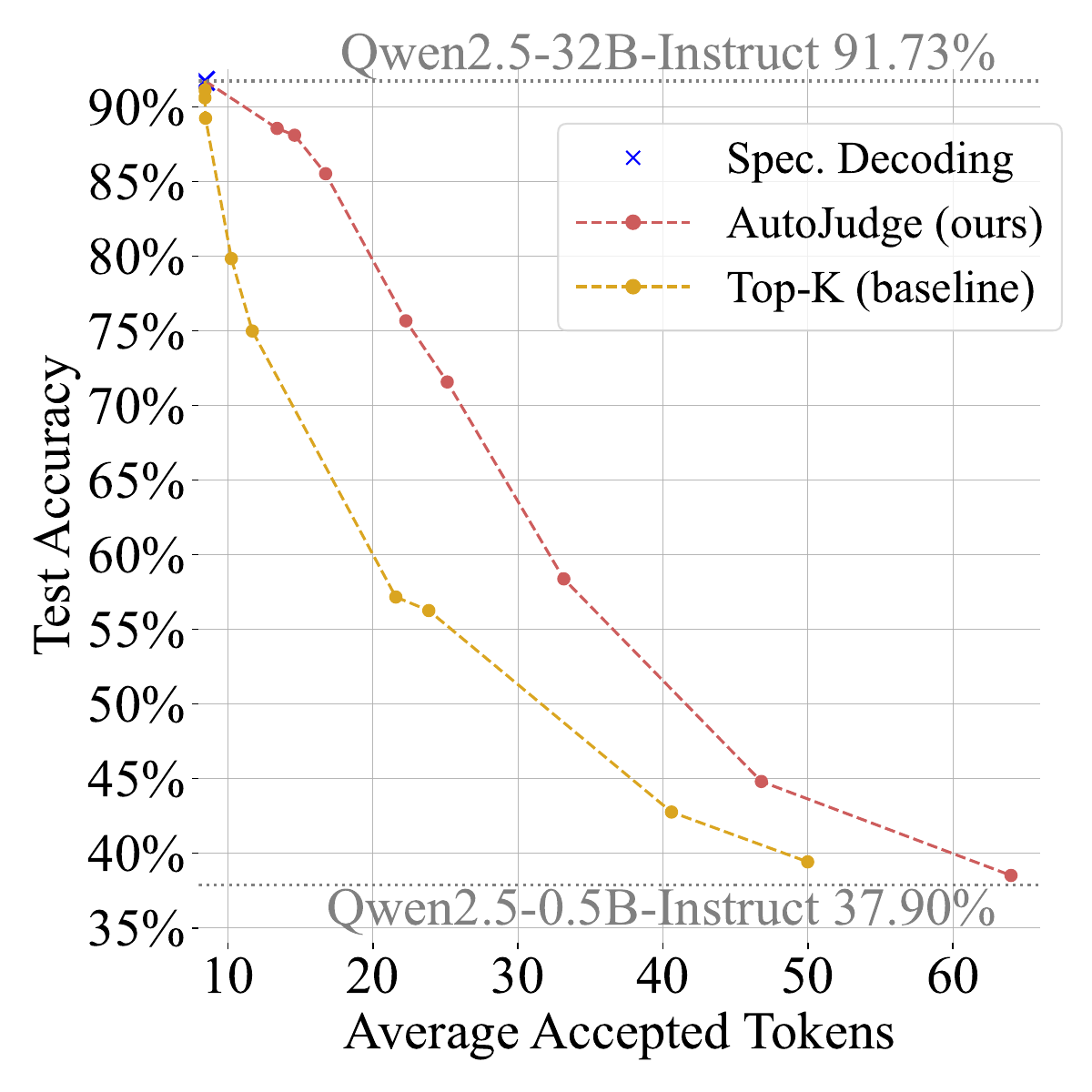}

    \caption{Accuracy and the average number of accepted tokens on GSM8K 8-shot for \textbf{(left)} Qwen2.5-0.5B draft / Qwen2.5-7B target and \textbf{(right)} Qwen2.5-0.5B draft / Qwen2.5-32B target (all Instruct).}
    \label{fig:plots_gsm8k_qwen}
    \vspace{5px}
\end{figure}

We report two main metrics: downstream accuracy and the number of accepted tokens per speculative decoding cycle. 
The accuracy is measured as the exact match rate for the final answer extracted from the response as per standard GSM8K evaluation protocol. In turn, we report the decoding speed in terms of the number of tokens accepted per target model forward pass, aiming to decouple our results from the specific hardware configuration.
We evaluate AutoJudge with different classifier thresholds, balancing between accuracy and speed.
Our baselines are traditional speculative decoding, decoding with the draft model, and a simpler lossy speculative decoding method. In the latter, we accept a mismatching draft token if it is within top-$K$ most likely tokens of the target model, similarly to how it is defined in~\cite{bachmann2025judgedecoding}. We report $K{=}2,4,8,\dots,|V|$ for different speed-accuracy tradeoffs: increasing $K$ results in more accepted tokens but reduces accuracy.

The results in Figure~\ref{fig:main_plots_gsm8k} demonstrate that AutoJudge decoding can achieve substantial speedups. 
Notably, our algorithm can accept over 40 tokens per target model forward pass in 8-shot evaluations for the 8B draft / 70B target model pair with a ${\le}1\%$ change in accuracy. 
Varying the classifier threshold allows us to achieve even greater speedups at the cost of several percentage points drop in accuracy. 
The heuristic-based top-$K$ baseline also achieves some speedups, but at the cost of a significantly higher accuracy drawdown. 
We report additional setups in        Appendix~\ref{app:experiments_gsm8k}.

\begin{figure}[t]
    \centering
    \includegraphics[width=0.49\linewidth]{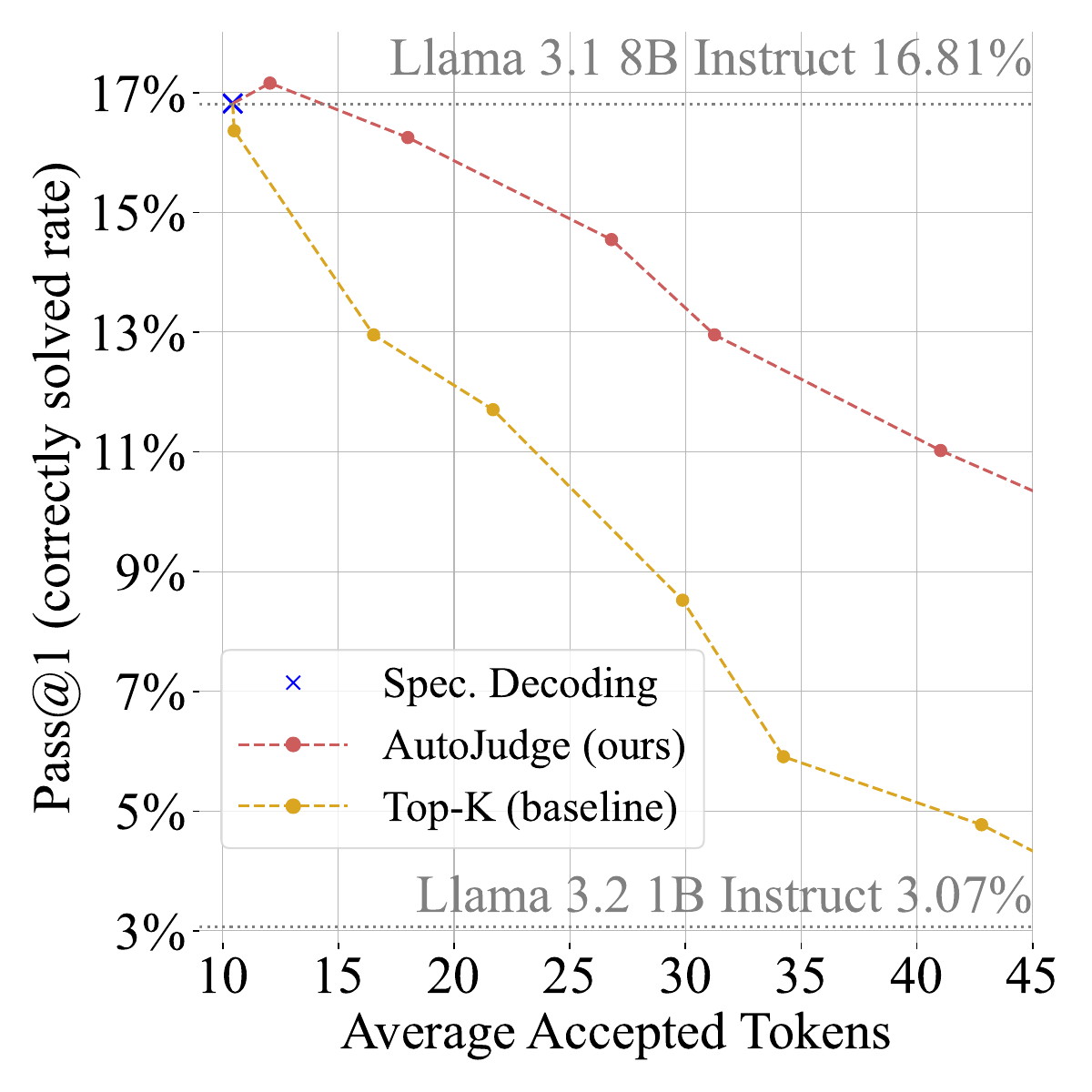}
    \includegraphics[width=0.49\linewidth]{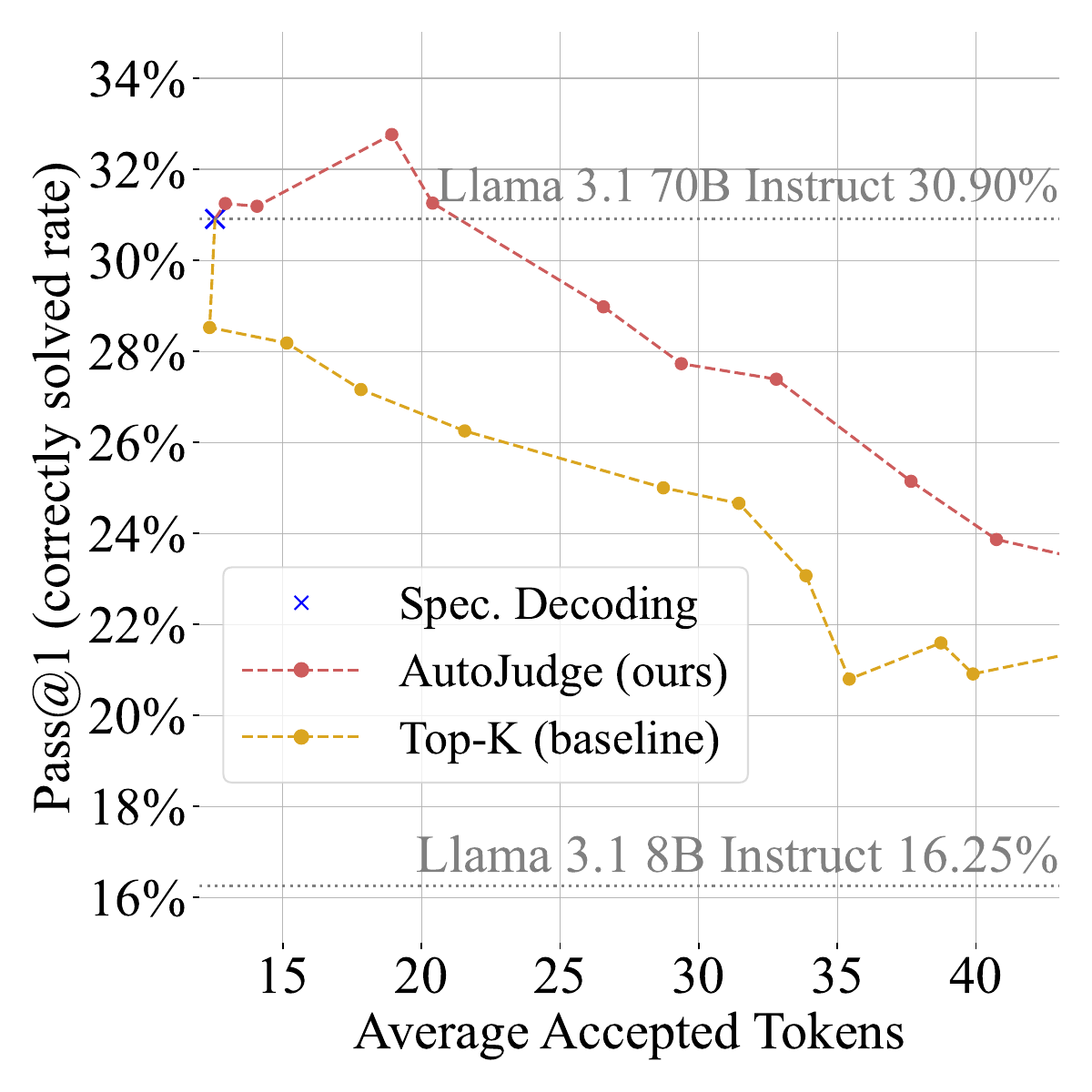}
    \vspace{-0px}
    \caption{
    Downstream Pass@1 and the average number of accepted tokens on LiveCodeBench for  \textbf{(left)} Llama-3.2 1B draft / Llama-3.1 8B target and \textbf{(right)} Llama 3.1 8B draft / Llama 3.1 70B target. We use Instruct versions for both model pairs and report additional details in Section~\ref{sect:exp_coding_lcb}.}
    \label{fig:main_results_lcb}
    \vspace{5px}
\end{figure}

\subsection{Programming with LiveCodeBench}\label{sect:exp_coding_lcb}
\vspace{-5px}

Next, we test if the AutoJudge search algorithm is able to generalize between domains.
For this purpose, we evaluate the same model family on the LiveCodeBench~\citep{jain2024livecodebenchholisticcontaminationfree} programming benchmark.
Here, we use the \texttt{code\_generation\_lite}\footnote{\href{https://huggingface.co/datasets/livecodebench/code_generation_lite}{\texttt{huggingface.co/datasets/livecodebench/code\_generation\_lite}}} dataset with the version tag \texttt{release\_v5}. 
The dataset contains 880 programming tasks; we evaluate on all three subsets: \texttt{easy}, \texttt{medium} and \texttt{hard}. 
Since LiveCodeBench does not have a dedicated training split, we evaluate using out-of-fold predictions. Namely, we split the dataset randomly into 5 folds. For each fold, we evaluate using the classifier trained on the 4 remaining folds. We use the standard evaluation protocol: extracting the generated code and evaluating it using the benchmark's built-in test suite. We follow the standard evaluation protocol for this benchmark and report Pass@1 in zero-shot setting.

Similarly to Section~\ref{sect:exp_math_gsm8k}, we use the training data to find important tokens --- this time, in terms of the resulting program correctness, measured as passing tests. Since the calibration dataset is smaller and further subdivided into folds, we only have ${\approx}$27K mismatching tokens to train the classifier (with a slight ${\le} 0.5K$ variation depending on the active fold). Furthermore, we found that only ${\approx}5\%$ of the mismatching tokens were deemed to affect the output quality. We provide example token assignments in Figure~\ref{fig:examples} (right): notably, the tokens deemed important in that case would not appear in GSM8K in the same context. 
We otherwise follow the same training and evaluation protocol as before.

The results in Figure~\ref{fig:main_results_lcb} are similar to what we observed in Section~\ref{sect:exp_math_gsm8k}: AutoJudge decoding can accept over 22 tokens per speculative decoding cycle at the cost of a ${\approx}2\%$ accuracy decrease. 
This results in approximately $2{\times}$ increase over traditional speculative decoding~\citep{leviathan2023fast}. 
The top-$K$ baseline can similarly achieve \textit{some} increase in the number of accepted tokens, but AutoJudge decoding offers significantly better quality-speed tradeoffs across all configurations.
We report additional configurations and threshold values in Appendix~\ref{app:experiments_lcb}, including the setup for AutoJudge decoding ``out-of-domain'', i.e. using the classifier trained on GSM8K for LiveCodeBench evaluation. This out-of-domain configuration results in inferior performance, which aligns with our hypothesis that the important tokens depend on the problem type and evaluation criteria.

\subsection{Inference Benchmarks with vLLM}\label{sect:exp_vllm_inference}
\vspace{-5px}

\begin{figure}[b]
  \vspace{10px}
  \captionof{table}{Inference speed benchmarks on GSM8K \underline{0-shot} with vLLM for \textbf{(left)} 1B draft / 8B target models with tuned window size (baseline {=} 8, AutoJudge {=} 10) and \textbf{(right)} for 8B draft / 70B target models (all Instruct) with tuned window size (baseline {=} 8, AutoJudge {=} 32 ).}
  \label{tab:main_vllm_0shot_1b_8b_and_8b_70b}
  \vspace{5px}
    \begin{minipage}[b]{.48\textwidth}
      \setlength{\tabcolsep}{4pt}
      \renewcommand{\arraystretch}{1.25}
      \begin{tabular}{l|cccc}
      \multicolumn{5}{l}{Llama 3.2 1B draft / 3.1 8B target (0-shot)}  \\
      \toprule
      \textbf{Threshold} & 0.06  & \textbf{0.12} & 0.15 & 0.16 \\
      \midrule
      Accuracy, \% & 83.1 & \textbf{80.2} & 79.8 & 77.4 \\
      Speed, tokens/s & 149.2 & \textbf{169.2} & 171.2 & 173.9 \\
      \midrule
      \multicolumn{4}{l}{\itshape Speculative Decoding:}  147.7 tokens/s \\
      \midrule
       Speedup(ours) & 1.01 & \textbf{1.14} & 1.15 & 1.17 \\
      \bottomrule
    \end{tabular}
  \end{minipage}
  \hspace{7px}
    \begin{minipage}[b]{.48\textwidth}
      \setlength{\tabcolsep}{4pt}
      \renewcommand{\arraystretch}{1.25}
\begin{tabular}{l|cccc}
\multicolumn{5}{l}{Llama 3.1 8B draft / 3.1 70B target (0-shot)}  \\
\toprule
\textbf{Threshold} & 0.005 & 0.031 & \textbf{0.145} & 0.230 \\
\midrule
Accuracy, \% & 92.0 & 91.9 & \textbf{89.9} & 88.0 \\
Speed, tokens/s & 72.3  & 80.6 & \textbf{107.4} & 109.6 \\
\midrule
\multicolumn{4}{l}{\itshape Speculative Decoding:} 72.3 tokens/s \\
\midrule
Speedup(ours) & 1.0 & 1.11 & \textbf{1.49} & 1.52 \\
\bottomrule
\end{tabular}
\end{minipage}
  \vspace{5px}
\end{figure}

In this section, we report the GPU inference speed of speculative decoding with AutoJudge classifiers. To benchmark real-world inference speed, we integrated AutoJudge into vLLM speculative decoding. We use the same GSM8K evaluation setup as in Section~\ref{sect:exp_math_gsm8k} and evaluate 3 model pairs from Llama 3.x family: 1B/8B,  8B/70B and 8B/405B draft/target respectively (all Instruct).

We compare AutoJudge decoding against lossless vLLM speculative decoding~\citep{leviathan2023fast}. To ensure fair comparison, we tune draft window size for AutoJudge and baseline independently (see Appendix~\ref{app:individually_tuned_eval}). In each configuration, we report the absolute speed in tokens per second (batch size 1), as well as the relative \textbf{speedup over speculative decoding}.
We run 1B/8B model pair on a single A100-SXM4-80GB GPU; 8B/70B on 4 A100-SXM4-80G GPUs in tensor-parallel mode. Finally, the 8B/405B runs on 8 H100-SXM5-80GB GPUs with the 405B model loaded in FP8 precision. Additionally, we consider a setup where 8B/70B model pair runs on a single A100-SXM4-80GB GPU with RAM offloading (see Appendix~\ref{app:inference_offloading}). We provide additional configuration details in Appendix~\ref{app:inference}.

\begin{figure}[t]
  \vspace{-10px}
  \captionof{table}{Inference speed with vLLM for \textbf{(left)} Llama 3.1 8B draft / 405B target models on GSM8K \underline{0-shot} with tuned window size (baseline=14, AutoJudge{=}20). \textbf{(right)} Llama 3.1 8B draft / 70B target (all Instruct) on GSM8K \underline{8-shot with offloading}, tuned window size (baseline{=}10, AutoJudge=48).}

  \label{tab:main_vllm_0shot_8b_405b_and_8b_70b_offloading}
  \vspace{5px}
    \begin{minipage}[b]{.48\textwidth}
      \setlength{\tabcolsep}{4pt}
      \renewcommand{\arraystretch}{1.25}
      \begin{tabular}{l|cccc}
        \multicolumn{5}{l}{Llama 3.1 8B draft / 3.1 405B target (0-shot)}  \\
        \toprule
        \textbf{Threshold} & 0.01 & \textbf{0.05} & 0.09 & 0.14 \\
        \midrule
        Accuracy, \% & 96.1 &\textbf{ 93.4} & 92.5 & 91.5 \\
        Speed, tokens/s & 50.6 & \textbf{58.0} & 58.5 & 60.1 \\
        \midrule
        \multicolumn{4}{l}{\itshape Speculative Decoding:} 50.0 tokens/s \\
        \midrule
        Speedup(ours) & 1.01 & \textbf{1.16 }& 1.17 & 1.20 \\
        \bottomrule
      \end{tabular}
    \end{minipage}
  \hspace{7px}
    \begin{minipage}[b]{.48\textwidth}
      \setlength{\tabcolsep}{4pt}
      \renewcommand{\arraystretch}{1.25}
      \begin{tabular}{l|cccc}
      \multicolumn{5}{l}{Llama 3.1 8B draft / 3.1 70B target (offload)}  \\
      \toprule
      \textbf{Threshold} & 0.03 & 0.05 & 0.11 & \textbf{0.28} \\
      \midrule
      Accuracy, \% & 95.4 & 94.8 & 93.4 & \textbf{90.4} \\
      Speed, tokens/s & 1.4 & 1.6 & 1.9 & \textbf{2.4} \\
      \midrule
      \multicolumn{4}{l}{\itshape Speculative Decoding:}  1.19 tokens/s \\
      \midrule
       Speedup(ours) & 1.20 & 1.31 & 1.59 & \textbf{1.98} \\
      \bottomrule
    \end{tabular}
  \end{minipage}
  \vspace{5px}
\end{figure}

Our results in Tables~\ref{tab:main_vllm_0shot_1b_8b_and_8b_70b} and~\ref{tab:main_vllm_0shot_8b_405b_and_8b_70b_offloading} demonstrate consistent improvements across all model pairs, with particularly high speedups for 8B/70B with and without offloading. This confirms that our earlier results in terms of accepted tokens (Sections~\ref{sect:exp_math_gsm8k}~\&~\ref{sect:exp_coding_lcb}) translate to real-world tokens per second.
Namely, if we allow $\approx 2\%$ change in accuracy, our vLLM inference with AutoJudge classifier can achieve $107.4$ tokens per second for 8B/70B model pair, which translates into about $1.5\times$ speedup compared to regular speculative decoding. This improvement is even more noticeable for hybrid setup with offloaded target model (in Table~\ref{tab:main_vllm_0shot_8b_405b_and_8b_70b_offloading}), where AutoJudge achieves up to $2\times$ speedup.

\paragraph{Additional evaluations.} To better explore the wide variety of tasks, methods and hardware setups, we report multiple series of additional experiments in supplementary materials. We evaluate AutoJudge decoding with EAGLE-2 in Appendix~\ref{app:eagle}. Next, we generalize to open-ended problems (question answering and creative writing) with LLM-as-a-judge evaluation in Appendix~\ref{app:exp_llmjudge_arena}. We also evaluate how AutoJudge classifiers transfer to adjacent problems in Appendix~\ref{app:math_transfer}. Finally, we benchmark with equal window size in Appendix~\ref{app:experiments} and control for vLLM nondeterminism in Appendix ~\ref{app:vllm_inconsistency}.

\paragraph{Limitations.} Our approach assumes that the downstream task has a way to determine whether or not two solutions are equivalent. While this is true for many tasks, we found that AutoJudge sometimes struggles in open-ended tasks with no clear criteria for correct answers. It would be interesting to explore alternative algorithm designs more suitable for open-ended problems. Additionally, running Algorithm~\ref{alg:important_tokens_mining} consumes compute resources, mostly LLM inference, in the form of local inference or API calls. We discuss this in more detail in Appendices~\ref{app:hardware} and~\ref{app:energy}.

\vspace{-7px}
\section{Conclusion}\label{sect:discussion}
\vspace{-7px}

In this work, we propose and evaluate a fully automated protocol for task-specific acceleration of speculative decoding. Our experiments show that a simple procedure can automatically determine which of the mismatching tokens in the LLM response affect the downstream quality on a variety of tasks. Our runtime benchmarks demonstrate significant speedups on top of already tuned speculative decoding and EAGLE algorithms with minimal inference code modification.
We hope that AutoJudge can facilitate the use of Judge Decoding across different task types, languages and modalities.

One promising direction for future research is to focus on open-ended problems such as instruction following or creative writing. While AutoJudge already demonstrates some speedups on open-ended tasks, it was primarily designed for technical problems such as math and programming. It would be interesting to investigate how it can be extended for longform generations with noisy and informal quality criteria.
Another possible direction is to combine AutoJudge with additional speculative decoding algorithms, such as speculative decoding with tree-based drafts~\citep{specinfer,chen2024sequoiascalablerobusthardwareaware,specexec} or learned drafting heads~\citep{cai2024medusa,li2025eagle3scalinginferenceacceleration}.




\section*{Acknowledgements}

We would like to express our sincere gratitude to Denis Mazur for his valuable contributions to the implementation of API calls used in Algorithm~\ref{alg:important_tokens_mining} and for supporting inference with the Llama 405B model. We are also thankful for his positive influence on the overall atmosphere and team morale throughout the course of this project.

\bibliography{main}
\bibliographystyle{plainnat}

\newpage

\newpage
\appendix
\section{Additional Considerations for Section~\ref{sect:method_mining}}\label{app:algo_details}

\paragraph{Generalization to sampling.} In Section~\ref{sect:method_mining}, we assume that the generation procedure is deterministic, i.e. that the model performs ``greedy inference''. In practice, however, many applications work better with stochastic sampling~\citep{holtzman_nucleus_sampling}. However, this has an obvious caveat for Algorithm~\ref{alg:important_tokens_mining}: if the text generation process is stochastic, a token can be deemed important based not on its actual impact on the model outputs, but on the randomness of the decoding procedure.

To generalize our approach for stochastic generation, we take advantage of the well-known Gumbel-max trick~\citep{gumbel1954statistical}. To recap, if we add independent Gumbel-distributed random variables to each predicted logit and take the index of the maximum, the probability that a certain index will be chosen is equal to the softmax of the original logits.

In case of Algorithm~\ref{alg:important_tokens_mining}, we use the Gumbel-max trick to reparameterize stochastic sampling from the model as deterministic sampling conditioned on a predefined random state $s \gets \textsc{randBits}(N)$. Given a prompt $x$, a response prefix $y_{1:t}$ and model parameters $\theta$, we generate the next token as follows:

$$y_{next} = \underset{i}{\arg\max} \log P(i | x \oplus y_{1:t}, \theta) + \textsc{GumbelPRNG}(s \oplus x \oplus y_{1:t}),$$

where \textsc{GumbelPRNG} is a function that samples a pseudo-random variable from standard Gumbel distribution based on an input seed $s \oplus x \oplus y_{1:t}$. To recall, $\oplus$ denotes concatenation. This way, $y_{next}$ is distributed as $P(y_{next} | x \oplus y_{1:t}, \theta)$, but it is deterministic when conditioned on the random state $s$. Hence, we sample a random state $s$ once at the beginning of Algorithm\ref{alg:important_tokens_mining}, the entire procedure after that will also be conditionally deterministic (given $s$).

\paragraph{Issues with na\"ive important token mining.} As we described earlier, Algorithm~\ref{alg:important_tokens_mining} is inherently sequential because it searches not for individual important tokens, but for important token combinations. In principle, it is tempting to consider a simpler algorithm that considers each token replacement in isolation and can run in parallel. However, when considering $\texttt{[target\_model\_gen\_0, draft\_token, target\_model\_gen\_1]}$ sequences only, a sufficiently strong target model might recover from even a low-quality token and still produce the correct answer. This results in a failure mode where all tokens are individually unimportant, but when all such tokens are \textit{jointly} replaced with their draft versions, the model fails to produce the correct answer. In our preliminary experiments, when using Llama-3.1-70B-Instruct~\cite{touvron2023llama} as the target model and Llama-3.2-1B-Instruct as the draft model, fewer than 1\% of the tokens were labeled as important with this simplified algorithm, whereas our main Algorithm~\ref{alg:important_tokens_mining} found substantially more. 
One interesting guarantee of Algorithm~\ref{alg:important_tokens_mining} over its na\"ive counterpart is that whenever draft and target models produce different (non-equivalent) answers to a given prompt, our algorithm will find at least one important token, whereas the na\"ive algorithm may find none.

\paragraph{On starting conditions for the important token search.} To recall, mining important tokens can be viewed as a shortest path search algorithm in a tree of possible mismatch choices. When performing this type of search, there are two possible directions that one can search from. In Algorithm~\ref{alg:important_tokens_mining}, we start from the target model outputs and iteratively (greedily) replace the mismatching tokens with their draft versions. However, one could also start from the draft model outputs and iteratively swap in target model outputs until the answer becomes equivalent to that of the target model. If we were to use an exhaustive search algorithm, both approaches would converge to the same important token labeling. However, since we are using a semi-greedy algorithm, it is easier to start with an already correct solution and simplify it, as opposed to starting with a wrong one and attempting to fix it.

\section{Additional Details on Classifier Training}\label{app:classifier_inputs}

As we discussed earlier in Section~\ref{sect:method_train_head}, there are several important design choices that can affect the performance of an important token classifier in our setting. Here, we report the experiments that led us to use a linear classifier based on draft token embeddings encoded with both $\theta_{draft}$ and $\theta_{main}$. To that end, we compare the different classifier variants using the important token embeddings from the GSM8K~\citep{cobbe2021gsm8k} training subset (see Section~\ref{sect:exp_math_gsm8k}).

To compare different classifier configurations, we further divide the GSM8K training set into classifier training (90\%) and validation (10\%) subsets. We perform this division at the sample level, i.e., all labeled tokens from a given GSM8K sample are used either entirely for classifier training, or entirely for validation. We use the same training and validation subsets throughout this section.

\begin{figure}[ht]
    \centering
    \includegraphics[width=0.495\linewidth]{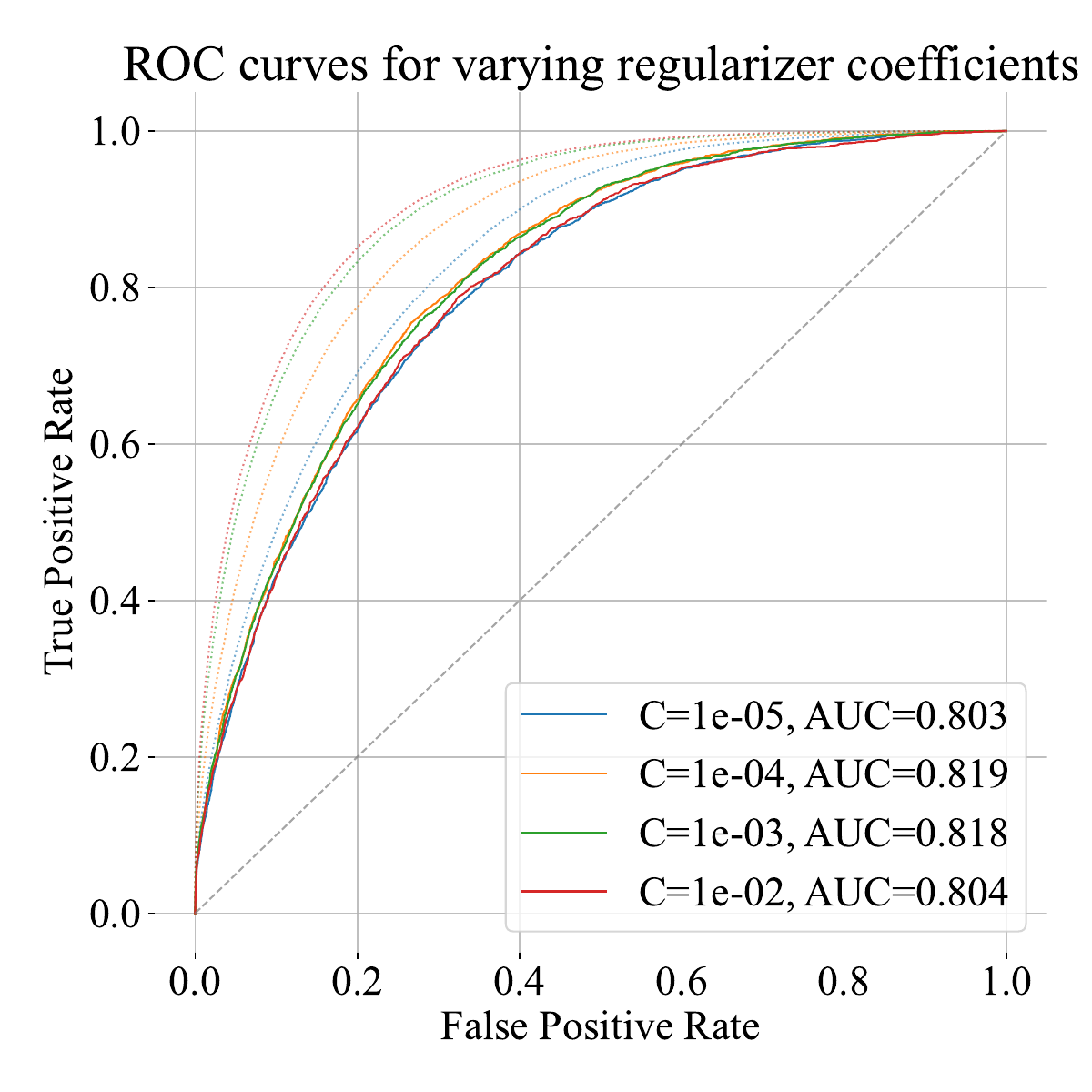}
    \includegraphics[width=0.495\linewidth]{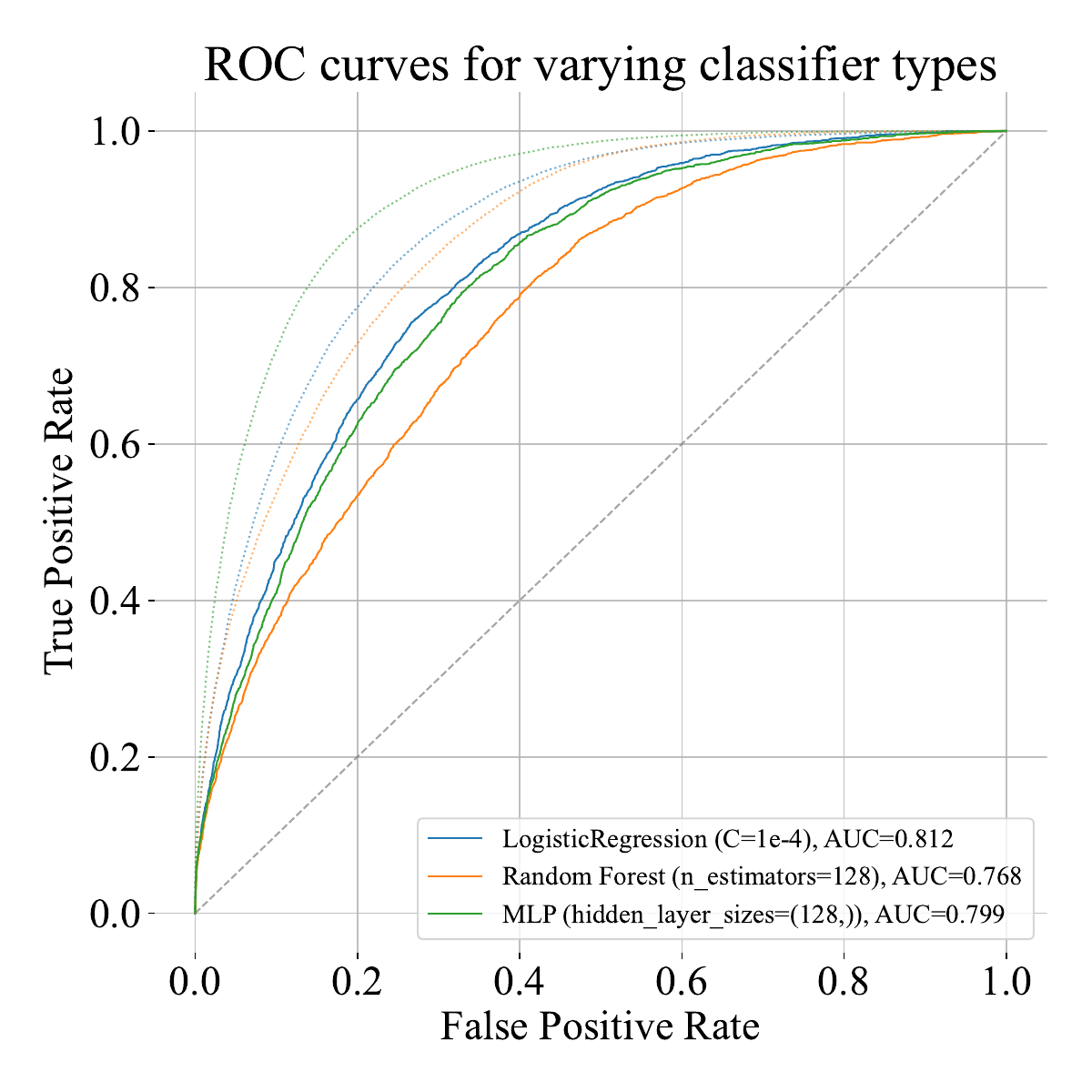}
    \caption{Receiver Operating Characteristics and the corresponding AUC values values for different Logistic Regression regularizers (left) and classifier types (right). Bold lines are validation curves and the dotted lines represent training curves. The AUC is reported in the legend (bottom right).}
    \label{fig:appendix_classifier_regularizer_and_architecture}
\end{figure}

For the first set of experiments (Figure~\ref{fig:appendix_classifier_regularizer_and_architecture}), we compare regularizer coefficients for Logistic Regression (left). We also report different classifier types: Logistic Regression, a Random Forest with 128 trees and a multi-layer perceptron (MLP) with a single hidden layer consisting of 128 hidden units with ReLU activation. For consistency, we run all models using scikit-learn~\citep{pedregosa2011scikit} v1.4.2 with all other settings kept to their default values. 
For LogisticRegression, we set \texttt{max\_iter} parameter to $500$. 
For MLP, we perform early stopping on yet another 10\% subset of the training set with the built-in default MLPClassifier early stopping parameters. 
For RandomForest, we use \texttt{min\_samples\_leaf} set to $0.001$. For this evaluation, all classifiers use draft and target model hidden states (concatenated) encoding the draft token, which is our main setup from Section~\ref{sect:method_train_head}.

The results in Figure~\ref{fig:appendix_classifier_regularizer_and_architecture} demonstrate that the classifier quality is fairly robust to the choice of the regularization hyperparameter. It is also fairly robust to the choice of the classifier architecture, barring perhaps the Random Forest classifier, which is overfitting the training data more than other models. Note that this does not necessarily mean that the MLP or tree-based classifiers are generally worse than linear models --- only that linear model is enough in our exact setup with a limited training set. We hypothesize that, if allowed to train on much larger dataset, the more complex models will be able to match and possibly outperform logistic regression.

Next, we compare classifier \textbf{inputs}. As we describe in Section~\ref{sect:method_train_head}, we use existing LLM hidden states from the last layer of $\theta_{draft}$ and $\theta_{target}$ since they are already computed during speculative decoding. This, however, leaves several possible choices about which hidden states should be used:\begin{itemize}[leftmargin=*]
    \item \textbf{Previous token embeddings}, last hidden states used to predict the mismatching token;
    \item \textbf{Draft token embeddings} are the next embeddings, obtained by encoding the draft token;
    \item \textbf{Target token embeddings} are the next embeddings, obtained by encoding the target token;
    \item \textbf{Both token embeddings} are concatenations of the draft and target token embeddings;
\end{itemize}

\begin{figure}[ht]
    \vspace{-15px}
    \centering
    \includegraphics[width=0.495\linewidth]{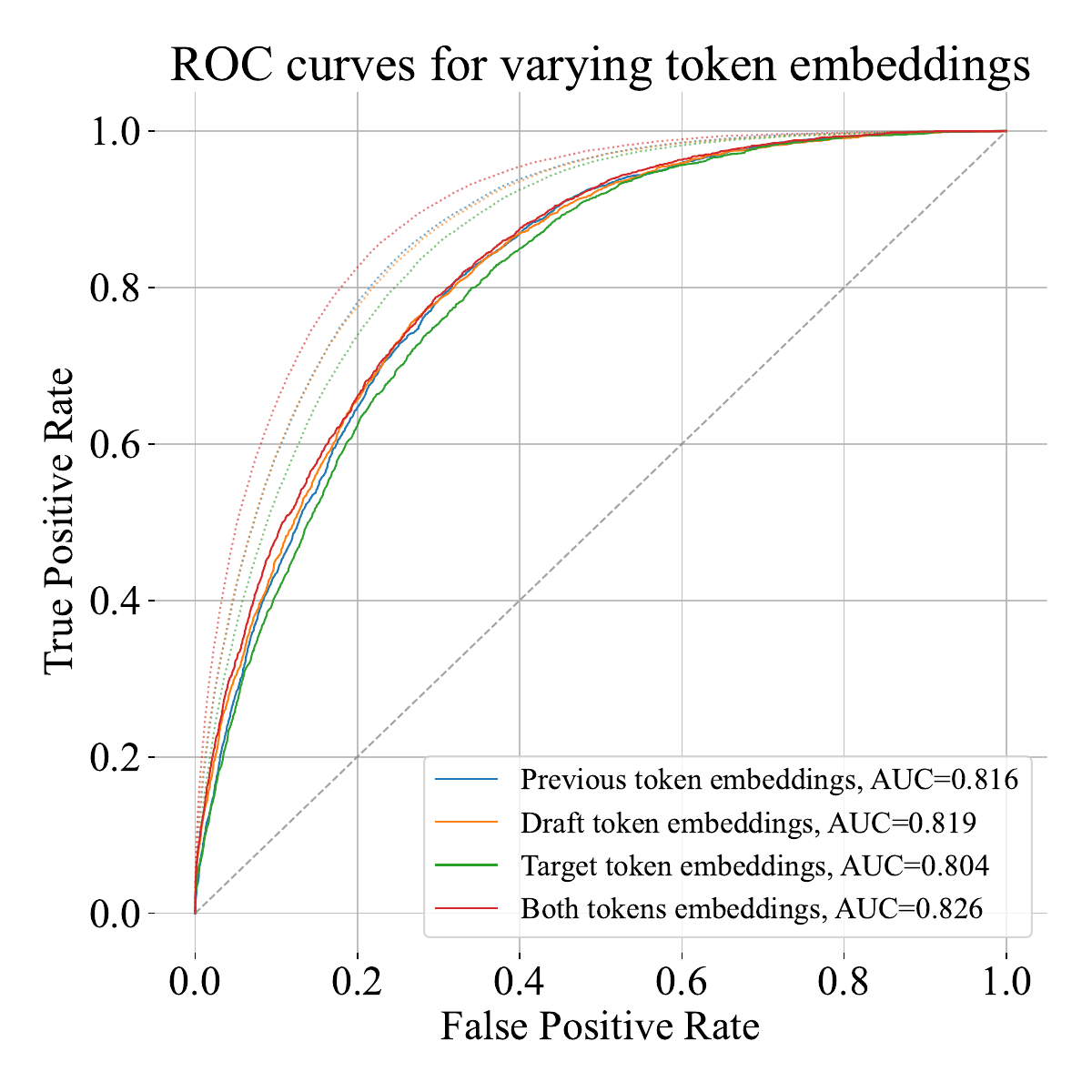}
    \includegraphics[width=0.495\linewidth]{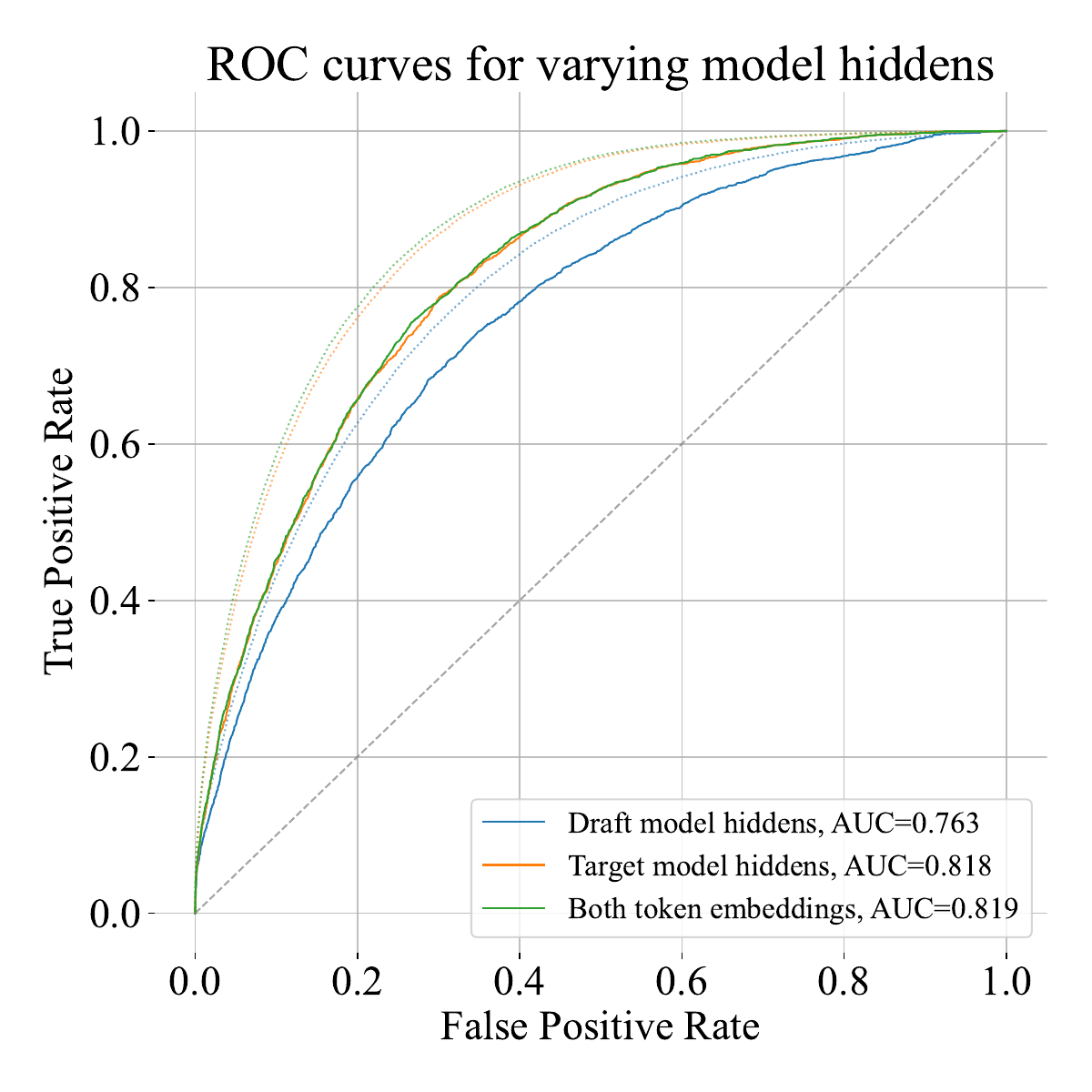}
    \caption{Receiver Operating Characteristics and the corresponding AUC scores (in the plot legend) for different classifier input tokens (left) and models (right). See Appendix~\ref{app:classifier_inputs} for details.}
    \label{fig:appendix_classifier_inputs}
\end{figure}

We compare the four input configurations in Figure~\ref{fig:appendix_classifier_inputs} (left), using Logistic Regression with $C{=}10^{-4}$ and both draft and target model hidden states (concatenated) for each case. The results suggest that a classifier that uses mismatching token embeddings (for draft \textit{or} target token) is significantly more accurate than using the preceding token embeddings (the ones used to predict the mismatch). In turn, using both token embeddings results in somewhat better performance than either of them. However, using both token embeddings introduces complications during inference time.

In normal speculative decoding, the algorithm already computes hidden states for draft tokens with both $\theta_{draft}$ (during draft generation) and $\theta_{target}$ (during verification). However, it does not compute embeddings for target tokens since those tokens are not known before the end of the verification stage --- and computing them already requires a forward pass with $\theta_{target}$. As a result, computing target (or both draft \& target) \textit{token} embeddings would require two sequential forward passes with $\theta_{target}$ --- one to determine the target tokens and detect mismatches, and the other to compute embeddings for those mismatching target tokens. In principle, one could devise a more sophisticated algorithm that computes only the $\theta_{draft}$ embeddings for mismatching target tokens or guesses the target tokens prior to the verification stage, but doing so would greatly complicate the implementation. Since the increase in the AUC score compared to using just the draft token embeddings is relatively small (Figure~\ref{fig:appendix_classifier_inputs}, on the left), we default to using draft token embeddings.

Additionally, we also test three model hidden states configurations for draft token embeddings: draft model hidden states, target model hidden states, and concatenated hidden states from both models in Figure~\ref{fig:appendix_classifier_inputs} (right). Here, using the target model hidden states results in superior accuracy to using the draft model. In turn, using both $\theta_{draft}$ and $\theta_{target}$ produces an additional, if marginal, increase in accuracy. However, since both hidden states are already available during inference, using them both does not pose additional complications. Though, some real world inference systems may make it more convenient to only use $\theta_{target}$ for classifier inputs since the AUC difference is within $1\%$.

\section{Precision Matters for Speculative Decoding}\label{app:precision}

When validating the AutoJudge algorithm, we found a peculiar implementation detail that can affect the real world performance of speculative decoding. Namely, \textit{when using the LLM in half precision, token embeddings can differ significantly (up to 10\%) between parallel and sequential forward passes on the same data.} In other words, if we record model hidden states as it generates a sequence, then encode the same sequence in parallel to recompute said hidden states, the two sets of hidden states will not match exactly. We attribute this to the fact that encoding tokens in parallel has a different summation order to encoding tokens one by one, which introduces small numeric errors. These errors compound over consecutive layers, resulting in larger errors in the final hidden states.

This is important for AutoJudge, since Algorithm~\ref{alg:important_tokens_mining} runs sequential inference with $\theta_{target}$ and parallel inference on $\theta_{draft}$, whereas inference-time speculative decoding does it the other way around: sequential calculations of $\theta_{draft}$ during the draft generation phase, then parallel forward pass with $\theta_{target}$ during the verification phase. As a result, the classifier is trained on features that can be significantly different from what they would be during inference. In contrast, running in full precision ($\texttt{float32}$) does not have such problems.

\begin{figure}[t]
    \centering
    \includegraphics[width=0.5\linewidth]{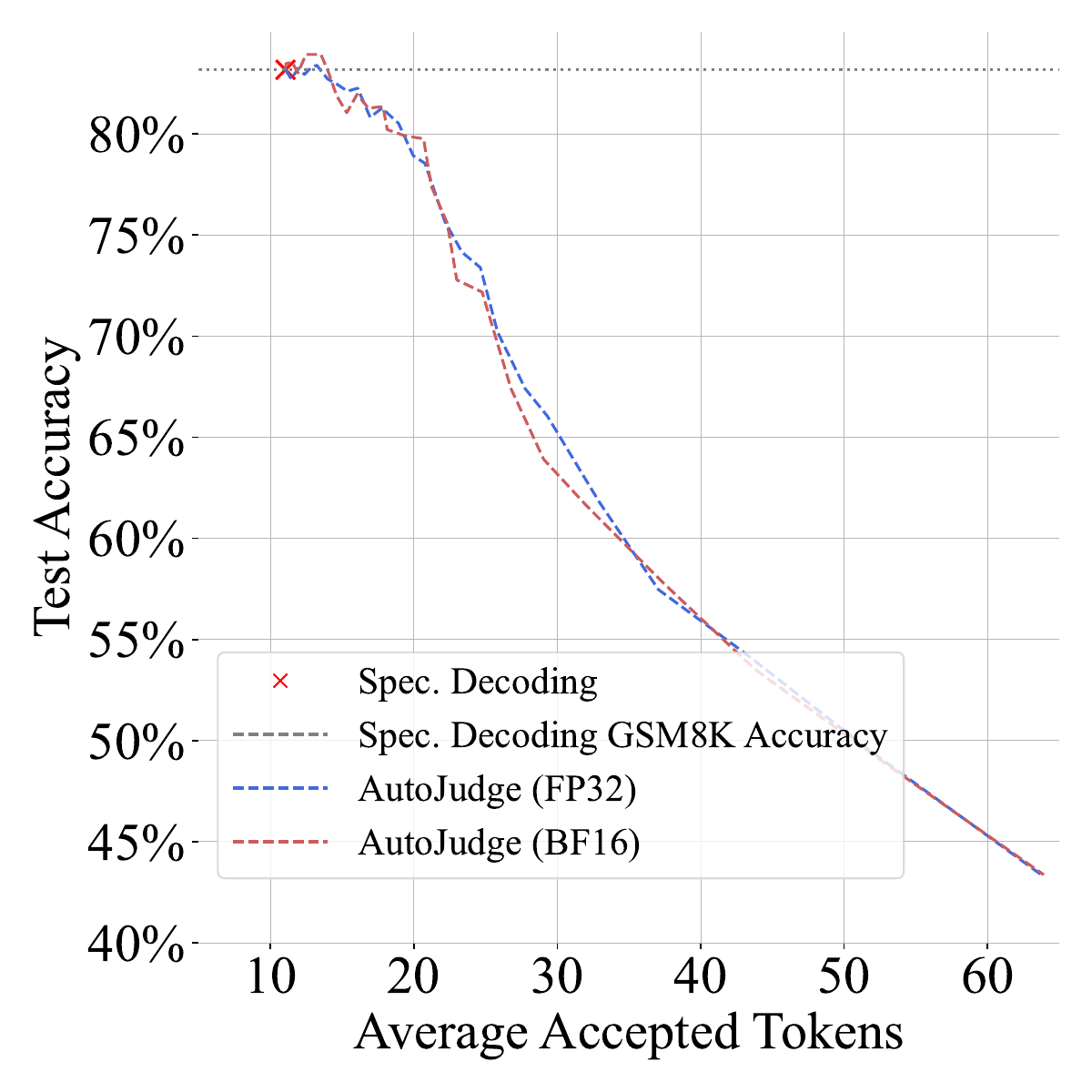}
    \caption{Accuracy on GSM8K and the number of accepted tokens per speculative decoding cycle in \texttt{float32} and \texttt{bfloat16} precision. The setup is the same as in Section~\ref{sect:exp_math_gsm8k}.}
    \label{fig:appendix_fp32_bf16_differences}
\end{figure}

In Figure~\ref{fig:appendix_fp32_bf16_differences}, we compare accuracy and acceptance rate trade-offs for different classifier thresholds in the same setup as in Section~\ref{sect:exp_math_gsm8k}. There are several ways to circumvent this problem. The most practical one would be to recompute target model embeddings for Algorithm~\ref{alg:important_tokens_mining} in a parallel forward pass and \textit{not} using the draft model embeddings (since adding them has negligible effect on accuracy, see Figure~\ref{fig:appendix_classifier_inputs}, right). As a result, the classifier would use $\theta_{target}$ embeddings computed in parallel over draft tokens during both training and inference.

\section{Additional Evaluations}\label{app:experiments}

\subsection{Additional Evaluations for Section~\ref{sect:exp_math_gsm8k}}\label{app:experiments_gsm8k}
\begin{figure}[h!]
    \centering
    \includegraphics[width=0.49\linewidth]{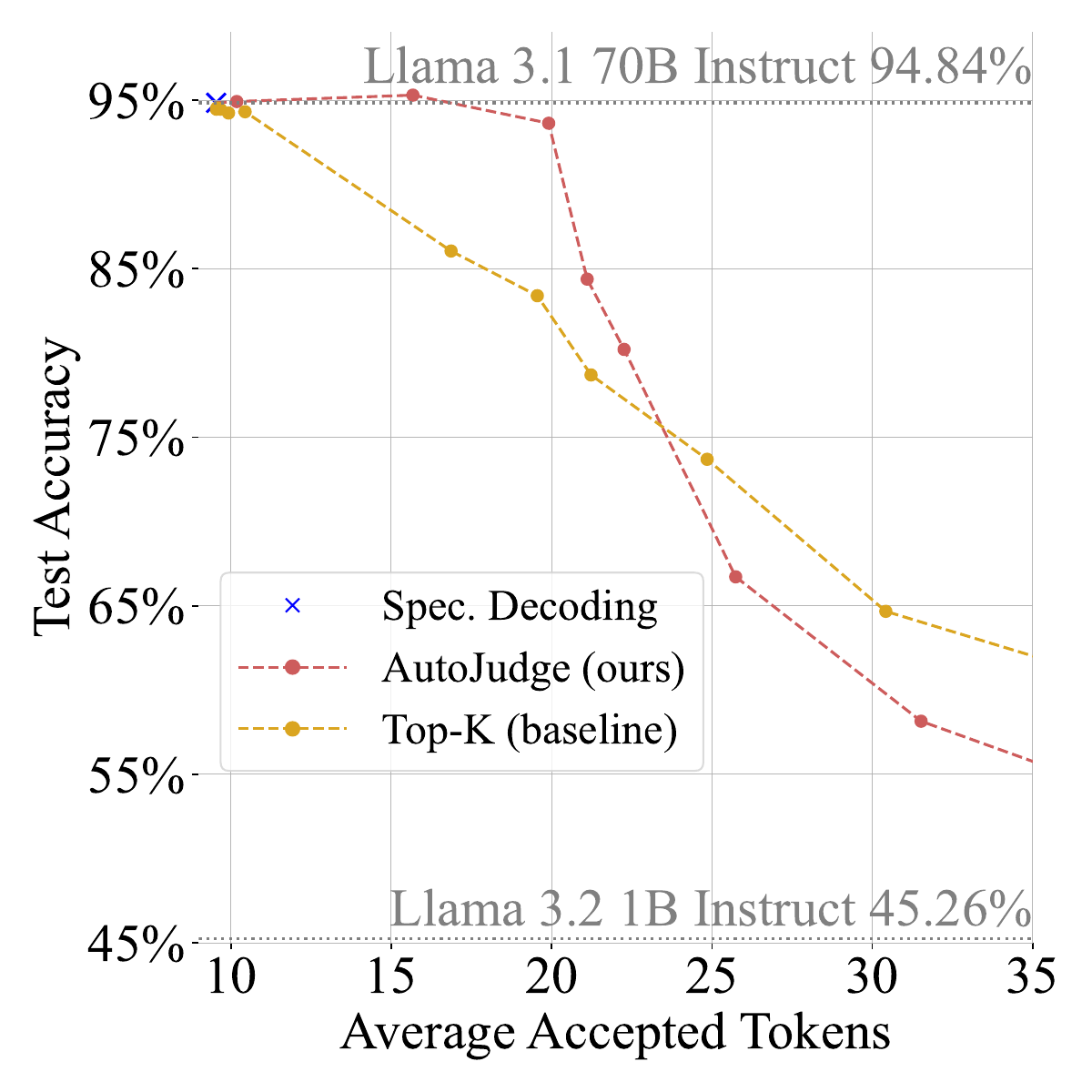} \includegraphics[width=0.49\linewidth]{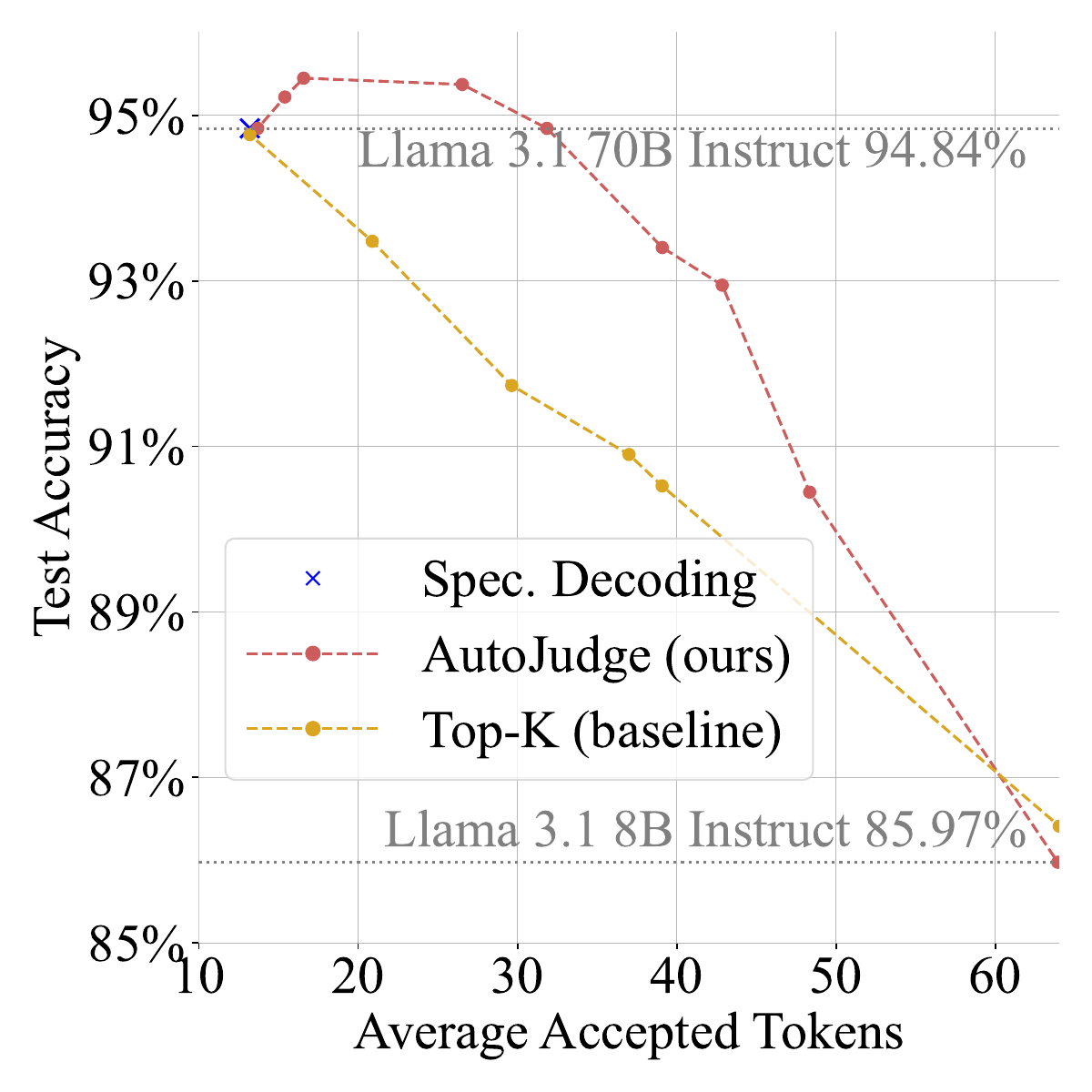}
    \vspace{-5px}
    \caption{Downstream accuracy and the average number of accepted tokens for GSM8K 8-shot with Llama-3.1-70B-Instruct target and Llama-3.2-1B-Instruct draft models (left) and Llama-3.1-70B-Instruct target and Llama-3.1-8B-Instruct draft models (right) }
    \label{fig:app_4_1_gsm_1b_8b_8shot_and_1b_70b_8shot}
\end{figure}

In Figure ~\ref{fig:app_4_1_gsm_1b_8b_8shot_and_1b_70b_8shot}, we report accuracy and the number of accepted tokens for  Llama-3.1-8B draft / Llama-3.1-70B target and Llama-3.1-8B draft / Llama-3.1-70B target model pairs in GSM8K 8-shot setup.


\begin{figure}[h]
    \centering
    \includegraphics[width=0.49\linewidth]{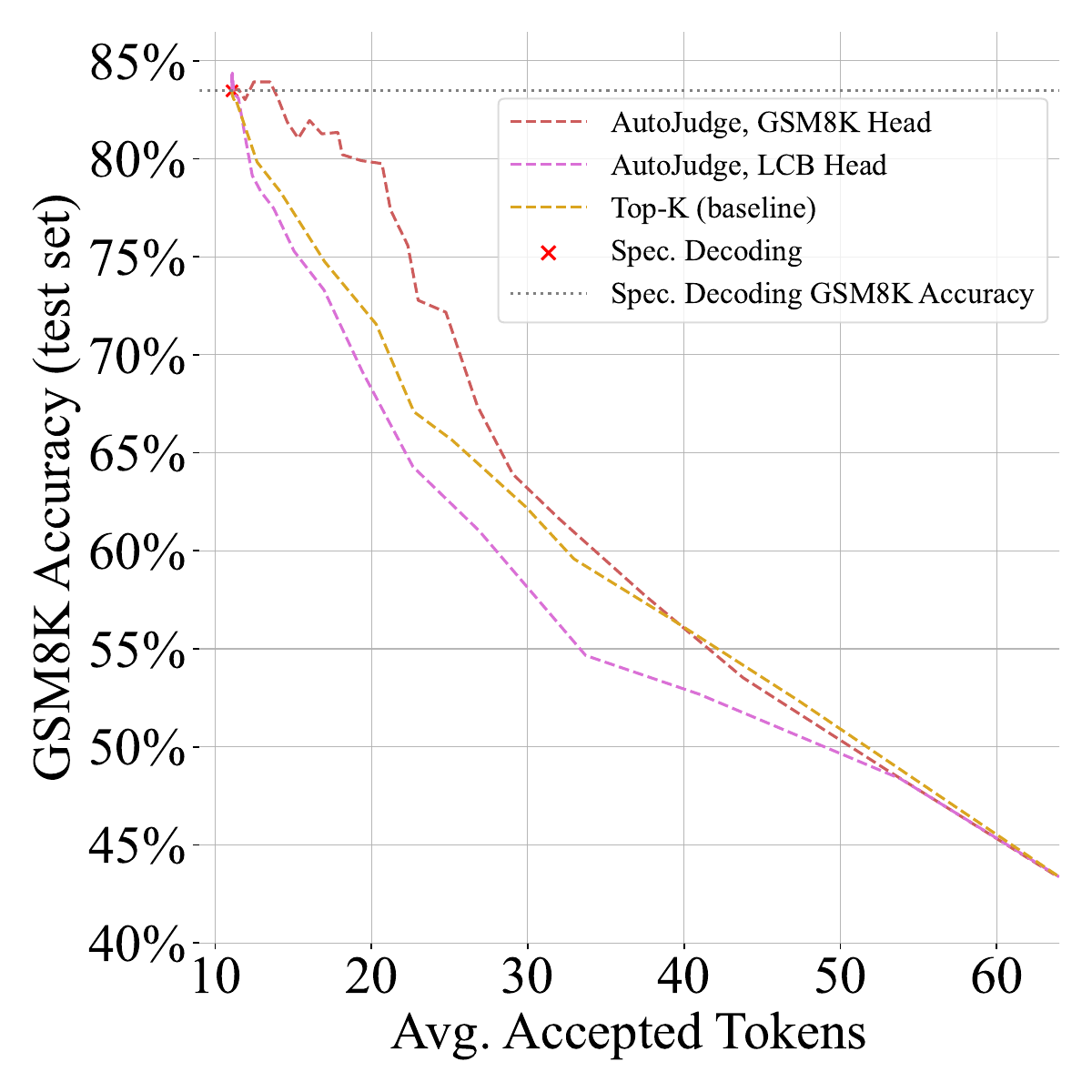} \includegraphics[width=0.49\linewidth]{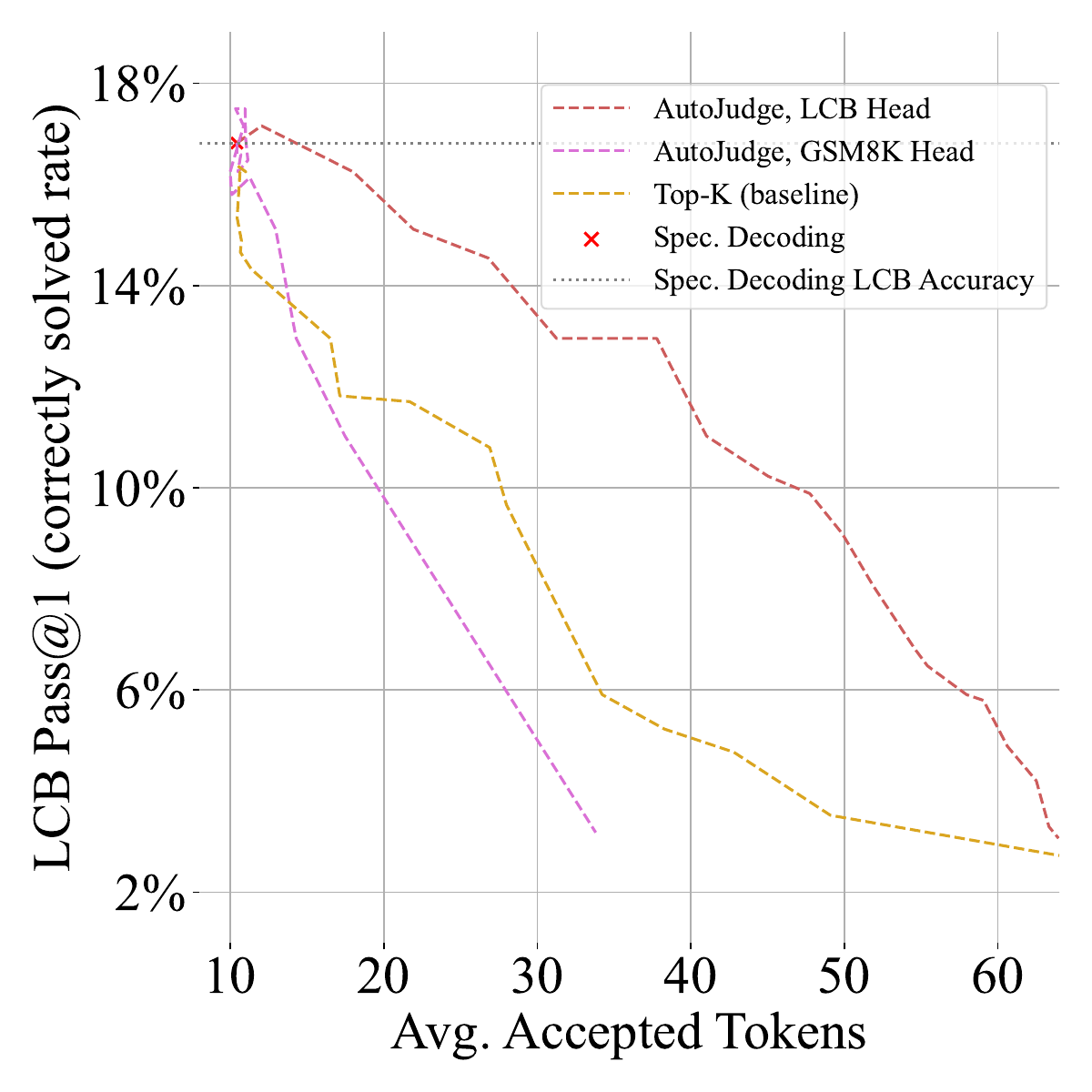}
    \vspace{-5px}
    \caption{Downstream accuracy and the average number of accepted tokens for GSM8K (left) and LiveCodeBench (right) with Llama-3.1-8B-Instruct target and Llama-3.2-1B-Instruct draft models.}
    \label{fig:app_4_1_gsm_lcb_1b_8b}
\end{figure}

\begin{figure}[h]
    \vspace{-10px}
    \centering
    \includegraphics[width=0.49\linewidth]{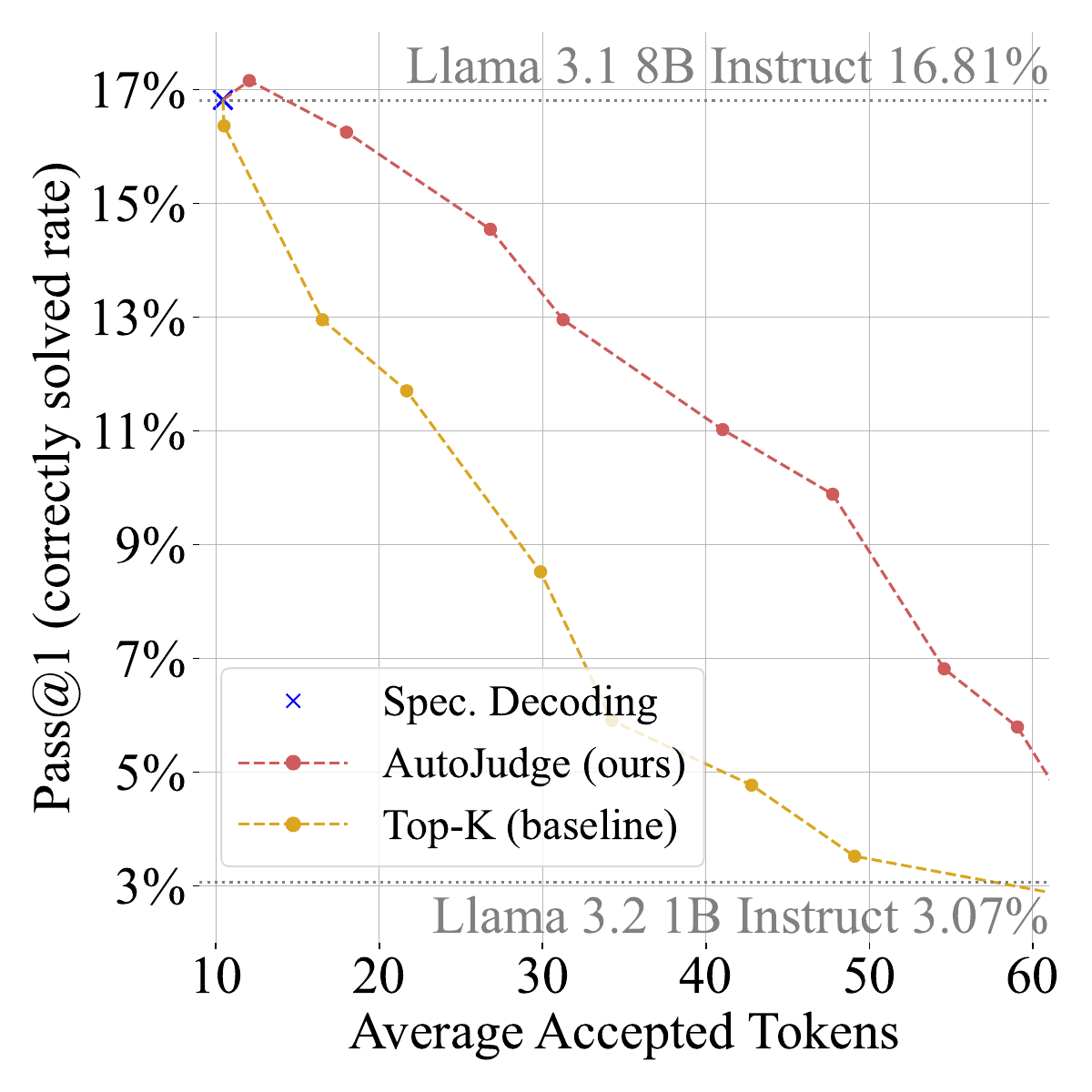} \includegraphics[width=0.49\linewidth]{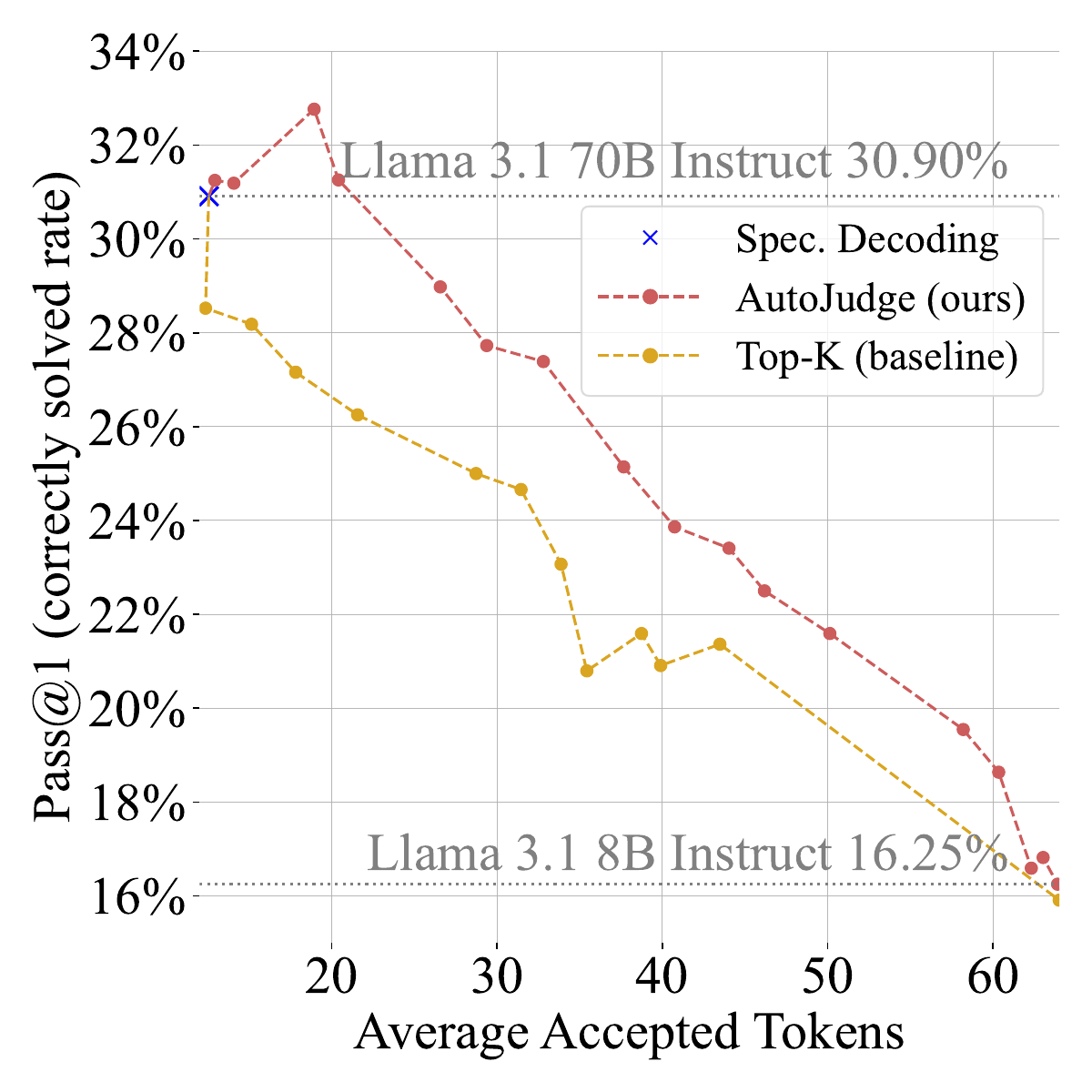}
    \vspace{-5px}
    \caption{Downstream accuracy and the average number of accepted tokens for LiveCodeBench with Llama-3.1-8B-Instruct target and Llama-3.2-1B-Instruct draft models (left) and Llama-3.1-70B-Instruct target and Llama-3.1-8B-Instruct draft models (right) }
    \label{fig:app_4_2_lcb_1b_8b_8shot_and_1b_70b_8shot}
\end{figure}

\subsection{Additional Evaluations for Sections~\ref{sect:exp_coding_lcb}}\label{app:experiments_lcb}

In this section, we provide additional classifier threshold evaluations for both model pairs in our LiveCodeBench setup. These results are reported in Figure~\ref{fig:app_4_2_lcb_1b_8b_8shot_and_1b_70b_8shot}, with 1B draft / 8B target models on the left and 8B draft / 70B target on the right. We use the same setup as in Section~\ref{sect:exp_coding_lcb} and the portion of the results at the top are exactly the same values that we reported in Figure~\ref{fig:main_results_lcb}.

Additionally, we evaluate the AutoJudge classifier trained on LiveCodeBench on GSM8K and vice versa to gauge the effect of task-specific training. We repost results for Llama-3.2-1B draft / Llama-3.1-8B target models pair in Figure~\ref{fig:app_4_1_gsm_lcb_1b_8b}. Predictably, these out-of-domain classifiers perform significantly worse. We attribute this to the fact that the GSM8K-trained classifier likely did not see any Python source code, whereas the LiveCodeBench classifier did not perform arithmetic operations and did not solve equations that are common in GSM8K.

\section{vLLM Inference Implementation Details}\label{app:inference}
To measure time efficiency of AutoJudge, we incorporate it into vLLM inference library.
The integration is built upon \verb|vllm==0.8.5|, \verb|torch==2.7.0| with CUDA 12.8 and \verb|transformers==4.51.3|.
We use a window size of $32$ and a batch size of $1$.
The implementation allows to perform batched inference without modifications, but we choose to evaluate with batch size $1$ for accurate measurements.
For efficiency, the current implementation predicts important tokens using only the hidden state of the target model, as we have shown that adding hidden states of the draft model does not substantially increase the accuracy.
We run evaluations for the 1B/8B model pair and for the 8B/70B pair 1 and 4 NVIDIA A100-SXM4-80GB GPUs respectively and on 8 NVIDIA H100-SXM5 80GB GPUs for 8B/405B in FP8 precision in the tensor parallel setup.

\section{Additional Inference Benchmarks: Equal Window Size}\label{app:inference_benchmarks}

In this section, we report additional runtime benchmarks that extend our evaluations from Section~\ref{sect:exp_vllm_inference}. For these experiments, we use a fixed window size of 32 for both vanilla speculative decoding and AutoJudge (as opposed to individually tuned window size in Section~\ref{sect:exp_vllm_inference}).

We report our results for GSM8K with 1B/8B model pair in Table~\ref{tab:additional_vllm_inference_1b_8b}, 8B/70B in Table~\ref{tab:additional_vllm_inference_8b_70b} and 1B/70B in Table \ref{tab:additional_vllm_inference_1b_70b}. All three model pairs show significant speedups relative to standard speculative decoding. Notably, our 8B/70B setup has up to 2.5x with about 2.5\% loss in accuracy.

\begin{figure}[h]
\vspace{-5px}
  \captionof{table}{Inference speed benchmarks on GSM8K for 1B draft / 8B target model on: (left) GSM8K 0-shot evaluation and (right) GSM8K 8-shot evaluation.}
  \label{tab:additional_vllm_inference_1b_8b}
  \centering
\begin{minipage}[b]{.48\textwidth}
  \setlength{\tabcolsep}{4pt}
  \renewcommand{\arraystretch}{1.25}
  \begin{tabular}{l|cccc}
    \multicolumn{5}{l}{Llama 3.2 1B draft / 3.1 8B target (0-shot)}  \\
    \toprule
    \textbf{Threshold} & 0.04  & 0.09 & 0.12 & \textbf{0.16} \\
    \midrule
    Accuracy, \% & 83.9 & 82.0 & 80.2 & \textbf{77.4} \\
    Speed, tokens/s & 89.1 & 106.8 & 121.0 & \textbf{146.2} \\
    \midrule
    \multicolumn{4}{l}{\itshape Speculative Decoding:} 84.5 tokens/s \\
    \midrule
    Speedup(ours) & 1.05 & 1.26 & 1.43 & \textbf{1.72} \\
    \bottomrule
  \end{tabular}
\end{minipage}
\hspace{7px}
\begin{minipage}[b]{.48\textwidth}
  \setlength{\tabcolsep}{4pt}
  \renewcommand{\arraystretch}{1.25}
\begin{tabular}{l|cccc}
\multicolumn{5}{l}{Llama 3.2 1B draft / 3.1 8B target (8-shot)}   \\
\toprule
\textbf{Threshold} & 0.050 & 0.087 & \textbf{0.098} & 0.133 \\
\midrule
Accuracy, \% & 85.7 & 85.1 & \textbf{84.5} & 83.1 \\
Speed, tokens/s & 66.5 & 74.3 & \textbf{76.3} & 83.3 \\
\midrule
\multicolumn{4}{l}{\itshape Speculative Decoding:} 62.4 tokens/s \\
\midrule
Speedup(ours) & 1.07 & 1.19 & \textbf{1.22} & 1.33 \\
\bottomrule
\end{tabular}\end{minipage}
  \vspace{5px}
\end{figure}

\begin{figure}[h]
  \captionof{table}{Inference speed benchmarks on GSM8K for 8B draft / 70B target model on: (left) GSM8K 0-shot evaluation and (right) GSM8K 8-shot evaluation.}
  \label{tab:additional_vllm_inference_8b_70b}
  \vspace{5px}
  \begin{minipage}[b]{.48\textwidth}
      \setlength{\tabcolsep}{4pt}
      \renewcommand{\arraystretch}{1.25}
\begin{tabular}{l|cccc}
\multicolumn{5}{l}{Llama 3.1 8B draft / 3.1 70B target (0-shot)}  \\
\toprule
\textbf{Threshold} & 0.005 & 0.031 & \textbf{0.145} & 0.230 \\
\midrule
Accuracy, \% & 92.0 & 91.9 & \textbf{89.9} & 88.0 \\
Speed, tokens/s & 41.8 & 80.6 & \textbf{107.4} & 109.5 \\
\midrule
\multicolumn{4}{l}{\itshape Speculative Decoding:} 40.6 tokens/s \\
\midrule
Speedup(ours) & 1.03 & 1.98 & \textbf{2.64 }& 2.70 \\
\bottomrule
\end{tabular}
\end{minipage}
  \hspace{7px}
  \begin{minipage}[b]{.48\textwidth}
      \renewcommand{\arraystretch}{1.25}
      \setlength{\tabcolsep}{4pt}
      \begin{tabular}{l|cccc}
        \multicolumn{5}{l}{Llama 3.1 8B draft / 3.1 70B target (8-shot)}  \\
        \toprule
        \textbf{Threshold} & 0.03 & 0.05 & \textbf{0.18} & 0.28 \\
        \midrule
        Accuracy: Accuracy, \% & 95.4 & 94.8 & \textbf{92.9} & 90.4 \\
        Speed, tokens/s & 64.4 & 71.1 &\textbf{ 79.3} & 86.2 \\
        \midrule
        \multicolumn{4}{l}{\itshape Speculative Decoding:} 40.5 tokens/s \\
        \midrule
        Speedup(ours) & 1.58 & 1.74 & \textbf{1.94} & 2.11 \\
        \bottomrule
        \end{tabular}
  \end{minipage}
  \vspace{5px}
\end{figure}

\begin{figure}[b]
  \captionof{table}{Inference speed benchmarks on GSM8K 8-shot with vLLM implementation for 1B draft / 70B target models.}
  \label{tab:additional_vllm_inference_1b_70b}
  \begin{minipage}[b]{.48\textwidth}
      \setlength{\tabcolsep}{4pt}
      \renewcommand{\arraystretch}{1.25}
      \begin{tabular}{l|cccc}
        \multicolumn{5}{l}{Llama 3.2 1B draft / 3.1 70B target (8-shot)} \\
        \toprule
        \textbf{Threshold}& 0.01 & 0.03 & \textbf{0.05} & 0.11 \\
        \midrule
        Accuracy, \% & 95.1 & 95.3 & \textbf{94.6} & 92.3\\
        Speed, tokens/s & 50.1 & 65.6 & \textbf{75.5 }& 79.9\\
        \midrule
        \multicolumn{4}{l}{\itshape Speculative Decoding:} 45.7 tokens/s \\
        \midrule
        Speedup(ours) & 1.10 & 1.44 & \textbf{1.65} & 1.75 \\
      \bottomrule
    \end{tabular}
  \end{minipage}
  \hspace{7px}

  \vspace{5px}
\end{figure}

\section{Evaluation with Individually Tuned Window Sizes} \label{app:individually_tuned_eval}

As we discussed in Section~\ref{sect:exp_vllm_inference}, we tune the speculation window size individually for AutoJudge and vanilla speculative decoding to provide a more competitive baseline. This is because traditional speculative decoding accepts, on average, less tokens and does not benefit from having a larger draft window size. Thus, the two algorithms often work best with different draft sizes.

In this section, we account for this by evaluating AutoJudge and speculative decoding with optimal window sizes for every model pair. We consider window sizes between 6 and 64 with the following values: $[6, 8, 10, 12, 14, 16, 20, 24, 26, 28, 32, 40, 48, 54, 64]$ and choose the best window size in terms of tokens per second. Note that this evaluation protocol suffers from the natural variance in latency (e.g. due to varying individual clock rates) since we report the highest value measured.

The results for 1B/8B models pair are reported in Table ~\ref{tab:best_vs_best_1b_8b_gsm_0_and_8_shots}. The results for 8B/70B pair are reported in Table~\ref{tab:best_vs_best_8b_70b_gsm_0_and_8_shots}. Overall, AutoJudge decoding still significantly outperforms standard speculative decoding.


\begin{figure}[h]

  \captionof{table}{\textbf{(left)} Inference speed benchmarks on GSM8K \underline{0-shot} with vLLM implementation for 1B draft / 8B target models with tuned window size (Spec. Decoding = 8 tokens, AutoJudge = 10 tokens) and \textbf{(right)} for 1B draft / 8B target models on GSM8K \underline{8-shot} with tuned window size (Spec. Decoding = 6 tokens, Autojudge = 10 tokens). }
  \label{tab:best_vs_best_1b_8b_gsm_0_and_8_shots}
  \vspace{5px}
    \begin{minipage}[b]{.48\textwidth}
      \setlength{\tabcolsep}{4pt}
      \renewcommand{\arraystretch}{1.25}
      \begin{tabular}{l|cccc}
      \multicolumn{5}{l}{Llama 3.2 1B draft / 3.1 8B target (0-shot)}  \\
      \toprule
      \textbf{Threshold} & 0.06  & \textbf{0.12} & 0.15 & 0.16 \\
      \midrule
      Accuracy, \% & 83.1 & \textbf{80.2} & 79.8 & 77.4 \\
      Speed, tokens/s & 149.2 & \textbf{169.2} & 171.2 & 173.9 \\
      \midrule
      \multicolumn{4}{l}{\itshape Speculative Decoding:}  147.7 tokens/s \\
      \midrule
       Speedup(ours) & 1.01 & \textbf{1.14} & 1.15 & 1.17 \\
      \bottomrule
    \end{tabular}
  \end{minipage}
  \hspace{7px}
    \begin{minipage}[b]{.48\textwidth}
      \setlength{\tabcolsep}{4pt}
      \renewcommand{\arraystretch}{1.25}
\begin{tabular}{l|cccc}
\multicolumn{5}{l}{Llama 3.2 1B draft / 3.1 8B target (8-shot)}  \\
\toprule
\textbf{Threshold} & 0.050 & \textbf{0.098} & 0.133 & 0.164 \\
\midrule
Accuracy, \% & 85.7 & \textbf{84.5} & 83.1 & 81.2 \\
Speed, tokens/s & 114.4  & \textbf{125.0 } & 132.1  & 139.3 \\
\midrule
\multicolumn{4}{l}{\itshape Speculative Decoding:} 116.8  tokens/s \\
\midrule
Speedup(ours) & 0.98 & \textbf{1.07 } & 1.13  & 1.19 \\
\bottomrule
\end{tabular}
\end{minipage}
  \vspace{5px}
\end{figure}

\begin{figure}[h]
  \captionof{table}{\textbf{(left)} Inference speed benchmarks on GSM8K \underline{0-shot} with vLLM implementation for 8B draft / 70B target models with tuned window size (Spec. Decoding = 8 tokens, Autojudge = 32 tokens) and \textbf{(right)} for 8B draft / 70B target models on GSM8K \underline{8-shot} with tuned window size (Spec. Decoding = 8 tokens, Autojudge = 16 tokens).}
  \label{tab:best_vs_best_8b_70b_gsm_0_and_8_shots}
  \vspace{5px}
    \begin{minipage}[b]{.48\textwidth}
      \setlength{\tabcolsep}{4pt}
      \renewcommand{\arraystretch}{1.25}
\begin{tabular}{l|cccc}
\multicolumn{5}{l}{Llama 3.1 8B draft / 3.1 70B target (0-shot)}  \\
\toprule
\textbf{Threshold} & 0.005 & 0.031 & \textbf{0.145} & 0.230 \\
\midrule
Accuracy, \% & 92.0 & 91.9 & \textbf{89.9} & 88.0 \\
Speed, tokens/s & 72.3 & 80.6 & \textbf{107.4} & 109.6 \\
\midrule
\multicolumn{4}{l}{\itshape Speculative Decoding:} 72.3 tokens/s \\
\midrule
Speedup(ours) & 1.0 & 1.11 & \textbf{1.49} & 1.52 \\
\bottomrule
\end{tabular}

\end{minipage}
  \hspace{7px}
    \begin{minipage}[b]{.48\textwidth}
      \setlength{\tabcolsep}{4pt}
      \renewcommand{\arraystretch}{1.25}
      \begin{tabular}{l|cccc}
      \multicolumn{5}{l}{Llama 3.1 8B draft / 3.1 70B target (8-shot)}  \\
      \toprule
      \textbf{Threshold} & 0.03 & 0.05 & \textbf{0.18} & 0.28 \\
      \midrule
        Accuracy, \% & 95.4 & 94.8 & \textbf{92.9} & 90.4 \\
      Speed, tokens/s & 69.0 & 73.1 & \textbf{83.6} & 84.5 \\
      \midrule
      \multicolumn{4}{l}{\itshape Speculative Decoding:}  57.3 tokens/s \\
      \midrule
       Speedup(ours) & 1.20 & 1.27 & \textbf{1.45} & 1.47 \\
      \bottomrule
      \end{tabular}
    \end{minipage}
  \vspace{5px}
\end{figure}

\section{Offloading}\label{app:inference_offloading}
As we presented in Section~\ref{sect:experiments}, AutoJudge can accept up to 40 tokens per verification cycle, which makes it naturally well-suited for scenarios with a large draft window. In offloading setups, the drafting step is significantly cheaper than verification, yet speculative decoding suffers from being able to accept only a few tokens on average. AutoJudge avoids this limitation by accepting substantially more tokens per verification cycle, leading to notable gains in offloading configurations.

As reported in Table~\ref{tab:best_vs_best_offloading}, the accuracy drop with AutoJudge does not exceed 3\% across thresholds, while achieving throughput between 1.4 and 2.4 tokens/s, compared to 1.19 tokens/s for speculative decoding. This corresponds to relative speedups of 1.2$\times$–1.98$\times$. 
It turns out that the optimal window size for AutoJudge is 48 tokens, whereas for speculative decoding it is only 8.
\begin{figure}[t!]
  \captionof{table}{Inference speed benchmarks on GSM8K 8-shot for 8B draft / 70B target models \textbf{with offloading} on a single NVIDIA A100-SXM4 GPU with tuned window size (Spec. Decoding = $10$ tokens, AutoJudge = $48$ tokens).}
  \label{tab:best_vs_best_offloading}
  \begin{center}
    \begin{minipage}[b]{.48\textwidth}
      \setlength{\tabcolsep}{4pt}
      \renewcommand{\arraystretch}{1.25}
      \begin{tabular}{l|cccc}
      \multicolumn{5}{l}{Llama 3.1 8B draft / 3.1 70B target (8-shot)}  \\
      \toprule
      \textbf{Threshold} & 0.03 & 0.05 & 0.11 & \textbf{0.28} \\
      \midrule
      Accuracy, \% & 95.4 & 94.8 & 93.4 & \textbf{90.4} \\
      peed, tokens/s & 1.4 & 1.6 & 1.9 & \textbf{2.4} \\
      \midrule
      \multicolumn{4}{l}{\itshape Speculative Decoding:}  1.19 tokens/s \\
      \midrule
       Speedup(ours) & 1.20 & 1.31 & 1.59 & \textbf{1.98} \\
      \bottomrule
    \end{tabular}
  \end{minipage}
  \end{center}
\end{figure}

\section{On the Instability of vLLM Inference for Accuracy Benchmarks} \label{app:vllm_inconsistency}
In Section~\ref{sect:experiments}, we evaluate AutoJudge decoding in terms of the number of accepted tokens per phase and the real-world tokens per second. For convenience, we only use vLLM implementation to evaluate the number of tokens per second and report all other metrics based on transformers / pytorch implementation.
This is because, in our preliminary experiments, we observed discrepancies between benchmark accuracy with vLLM inference and PyTorch. Upon further investigation, we found that vLLM inference sometimes produces inconsistent results. Namely, if we run vLLM \textbf{greedy} inference in \texttt{bfloat16} precision with speculative decoding, changing ``technical’’ hyperparameters such as window size can significantly affect the accuracy. To illustrate this, we measure GSM8K accuracy on 132 random test samples in the same setup as Section \ref{sect:exp_vllm_inference}. We use standard vLLM implementation of speculative decoding~\cite{leviathan2023fast} \textit{without any modifications to the vLLM codebase}.

\begin{figure}[h!]
  \captionof{table}{Llama 3.2 1B draft / 3.1 70B target, 10\% random test samples from GSM8K. Accuracy is measured based on vLLM generations with varied window size.}
  \label{tab:vllm_instability}
  \vspace{5px}
  \centering
\begin{tabular}{rr}
\toprule
Window Size & Accuracy, \% \\
\midrule
8 & 93.9 \\
10 & 93.9 \\
12 & 93.9 \\
14 & 93.9 \\
16 & 95.4 \\
20 & 95.4 \\
24 & 94.6 \\
26 & 94.6 \\
\bottomrule
\end{tabular}
\end{figure}

The results in Table ~\ref{tab:vllm_instability} illustrate the observed inconsistency: changing window size can affect the accuracy to a significant degree. This coincides with concurrent observations about inconsistent LLM inference from~\citet{yuan2025fp32deathchallengessolutions} and ~\citet{he2025nondeterminism}. To make our measurements more consistent, we report all accuracy and accepted token rates using a more stable implementation based on Hugging Face Transformers (included in our repository), only using vLLM to measure the speed in terms of tokens/second. 

\section{EAGLE experiments}\label{app:eagle}

Following the invention of speculative decoding, several lines of work proposed follow-up algorithms that incorporate trained speculation ``heads''\citep{cai2024medusa,li2024eagle,eagle2}, tree decoding~\citep{specinfer,chen2024sequoiascalablerobusthardwareaware,specexec} and many other improvements. 
In this section, we explore how AutoJudge generalizes to these more advanced decoding algorithms.

To that end, we integrate AutoJudge with the popular EAGLE-2 algorithm~\citep{eagle2}. Unlike the original speculative decoding, EAGLE-2 does not have a separate draft model, but trains a lightweight ``head'' to predict future tokens from target model hidden states. This allows EAGLE to draft tokens much faster, albeit less accurately than powerful standalone draft model.

We evaluate Llama 3.1 8B Instruct using the official pre-trained EAGLE heads\footnote{\url{https://huggingface.co/yuhuili/EAGLE-LLaMA3.1-Instruct-8B}}. Since there is no separate draft model, the classifier is trained  we only use target model hidden states when training. Additionally, since EAGLE draft model was not trained to produce long coherent drafts, we use a shorter draft window size of 8. We integrate AutoJudge with the official EAGLE implementation and set parameters that are compatible with the vLLM EAGLE implementation (\texttt{depth{=}window\_size{-}1}). Aside from that, we use the same evaluation protocol as in Section~\ref{sect:exp_math_gsm8k} for GSM8K.

\begin{figure}[ht]
    \centering
    \begin{minipage}{0.49\linewidth}
    \hspace{-10px}\includegraphics[height=95px]{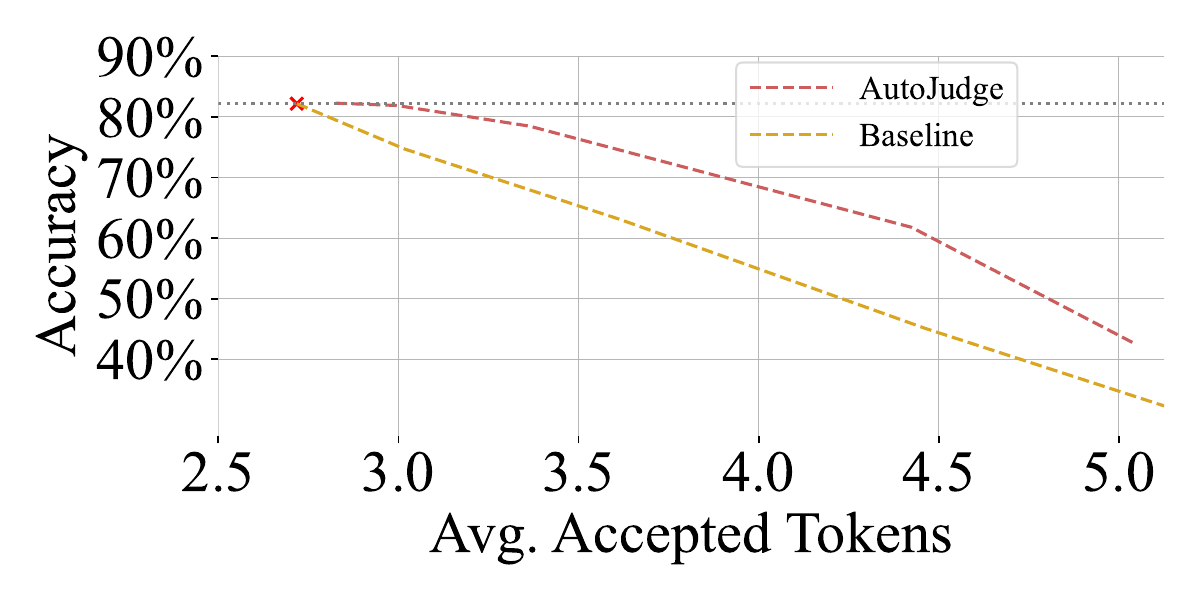}
    \vspace{-10px}
    \end{minipage}
    \hspace{-15px}
    \begin{minipage}{0.49\linewidth}
    \begin{tabular}{l|cccc}
    \multicolumn{5}{l}{Llama 3.1 8B Instruct + EAGLE head (0-shot)}  \\
    \toprule
    \textbf{Threshold} & 0.05 & 0.15 & \textbf{0.30} & 0.40 \\
    \midrule
    Accuracy, \%  & 81.4 & 81.3 & \textbf{81.0} & 78.1 \\
    Speed, tokens/s  & 91.1 & 96.8 & \textbf{102.6} & 107.5 \\
    \midrule
    \multicolumn{4}{l}{\itshape EAGLE:}  89.8 tokens/s \\
    \midrule
    Speedup(ours) & 1.01 & 1.08 & \textbf{1.14} & 1.20 \\
    \bottomrule
    \end{tabular}
    \end{minipage}
    \caption{Evaluating AutoJudge with EAGLE-2 draft head on GSM8K using Llama 3.1 8B Instruct model. (Left) Accuracy to accepted tokens in PyTorch, (right) vLLM inference speed on A100-80GB.}
    \label{fig:app_eagle}
\end{figure}

In Figure~\ref{fig:app_eagle}, we report GSM8K accuracy and the average number of accepted tokens for AutoJudge with the official PyTorch implementation\footnote{\url{https://github.com/SafeAILab/EAGLE/}} of EAGLE 2. We also report real-world inference speed (tokens per second) using the vLLM implementation on a single A100-80GB GPU. The results suggests that integrating AutoJudge with EAGLE can produce additional speedups on top of the highly efficient speculative decoding algorithm.

\section{Rule-Based Approaches Ablation}\label{app:rule_based_ablation}
\begin{figure}[ht]
    \captionof{table}{Comparison between AutoJudge and simple rule-based heuristic focused on mathematical tokens, GSM8K 0-shot, Llama 3.2 1B Instruct draft / 3.1 8B Instruct target.}
    \label{tab:rule_based_ablation}
    \centering
    \begin{tabular}{l|ccc}
    \toprule
    \textbf{Method} & \textbf{Math Only} & \textbf{Math \& Top-1024} & \textbf{Math \& Top-128} \\
    \midrule
    \textbf{Heuristic criterion} & & & \\
    \quad Accuracy, \% & 65.05 & 72.27 & 80.36 \\
    \quad Accepted Tokens & 39.7 & 22.3 & 15.5 \\
    \midrule
    \textbf{AutoJudge (nearest threshold)} & & & \\
    \quad Accuracy, \% & 57.8 & 75.6 & 81.0 \\
    \quad Accepted Tokens & 37.4 & 22.3 & 15.3 \\
    \bottomrule
    \end{tabular}
\end{figure}

To explore how AutoJudge decoding compares to rule-based methods for mathematical reasoning, we compare it against two heuristic alternatives. The first heuristic approach is to consider only the mathematical symbol tokens as ``important''. To that end, we filter the mismatching tokens that contain numbers, operations (e.g., + - * / =, etc.), as well as some common variables. Mismatches in these mathematical tokens are rejected, whereas mismatches in non-mathematical tokens are allowed. 
As it turns out (Table~\ref{tab:rule_based_ablation}), this algorithm misses important planning and logical steps that do not contain computations explicitly, which results in poor accuracy.
To address this, we introduce a second, more complex heuristic approach that combines the mathematical rule with the Top-K baseline we use in Section 4. This works somewhat better, but still does not outperform the learned AutoJudge classifier.

\section{Open-Ended Generation with LLM-as-a-Judge}\label{app:exp_llmjudge_arena}

As we discuss earlier in Section~\ref{sect:method_mining}, the choice of important tokens largely depends on what counts as an ``equivalent answer quality''. For mathematical reasoning, we can check whether the alternative response leads to the same numerical answer (up to notation) or an equivalent formula. For programming, we compare how the two programs behave in testing. However, not all LLM tasks have formal quality criteria. Open-ended tasks like creative writing and question answering have implicit quality criteria that are difficult to evaluate.
In this section, we investigate how AutoJudge generalizes to two open-ended problems: generative question answering and creative writing.

\textbf{Case A: Question Answering.} For this task, we evaluate Llama 3.x LLMs generative question answering on questions from the TriviaQA dataset~\citep{joshi-etal-2017-triviaqa}. We use the ``closed book'' setup, where the LLM receives only the question itself as the prompt, without any additional information (i.e. no search results). Then, we use a more powerful LLM-as-a-judge to compare responses against target model outputs. We mine important tokens on 500 training samples and evaluate on a subset of 100 validation samples. Since this dataset was intended for short answers (typically 1 sentence), we stop generation on \texttt{\textbackslash n} or after generating 120 tokens.

\textbf{Case B: Creative Writing.} We use the ``Creative Writing'' subset of Arena-Hard-Auto-v2.0~\citep{arenahard2024,li2024crowdsourced} that contains 250 creative writing tasks sourced from Chatbot Arena. These tasks include requests to write a poem on a certain topic, an imaginary dialogue transcript, or similar, in several languages. These tasks are even harder to judge than question answering, often boiling down to personal preference. We set 100 queries aside for evaluations and use the rest to mine important tokens, using the standard generation prompt from Arena-Hard-Auto-v2.

\textbf{LLM-as-a-judge.} We evaluate generations using a pairwise LLM-as-a-judge protocol inspired by Arena-Hard-Auto-v2.0. Under this protocol, the LLM ``judge'' does not test model answers against a pre-defined ``correct'' response, but compares two generations against each other. For our speculative decoding setup, we ask the LLM judge to compare target model generation against the output of speculative decoding with AutoJudge head. When comparing two generations, the LLM judge sees both responses A and B and chooses one of 5 options:\begin{enumerate}
    \item Assistant A is significantly better: [[A>\,\!>B]]
    \item Assistant A is slightly better: [[A>B]]
    \item Tie, relatively the same: [[A=B]]
    \item Assistant B is slightly better: [[B>A]]
    \item Assistant B is significantly better: [[B>\,\!>A]
\end{enumerate}

When evaluating on creative writing, we found that the default LLM judge often produces different answers between consecutive runs, both with sampling and greedy decoding, likely due to numerical instability~\citep{yuan2025fp32deathchallengessolutions}). To make our results more reliable we switch to a stronger \texttt{claude-sonnet-4-20250514} judge model\nocite{AnthropicClaude3.7Sonnet} and run it 3 times with majority voting:\begin{itemize}
    \item If there are more votes [[A>\,\!>B]] \& [[A>B]] than the opposite, then A is better.
    \item If there are more votes [[B>\,\!>A] \& [[B>A]]) than the opposite, then B is better.
    \item If there are equal number of votes favoring A and B or all votes are [[A=B]], we declare A and B are equivalent. Note that this rule does not differentiate between [[A>B]] and [[A>\,\!B]] as we found them to be uninformative.
\end{itemize}

\textbf{Mining important tokens.} We mine important tokens using Algorithm~\ref{alg:important_tokens_mining} with one change: in L12 (\textbf{if} $a \equiv \hat a$ \textbf{then} \dots), we ask the LLM judge to compare the alternative response $\hat y$ (with the draft token) against $y$. If the new response is rated strictly worse, we label the corresponding draft token as important and roll back to the target token. Otherwise (better or equal), we keep the draft token and label it unimportant, same as in the original algorithm. Note that the LLM judge compares \textbf{not} against the original target model response (L4 Alg.~\ref{alg:important_tokens_mining}), but against the current running response $y$ that contains unimportant draft tokens from previous iterations (L14). This change would have no effect for tasks with formal $a\equiv\hat a$ criteria from math and programming benchmarks due to transitivity. However, for LLM-as-a-judge, this results in more accurate labels and improves classifier accuracy. We only use this for training, not evaluation.

\begin{figure}[t]
    \centering
    \includegraphics[width=0.49\linewidth]{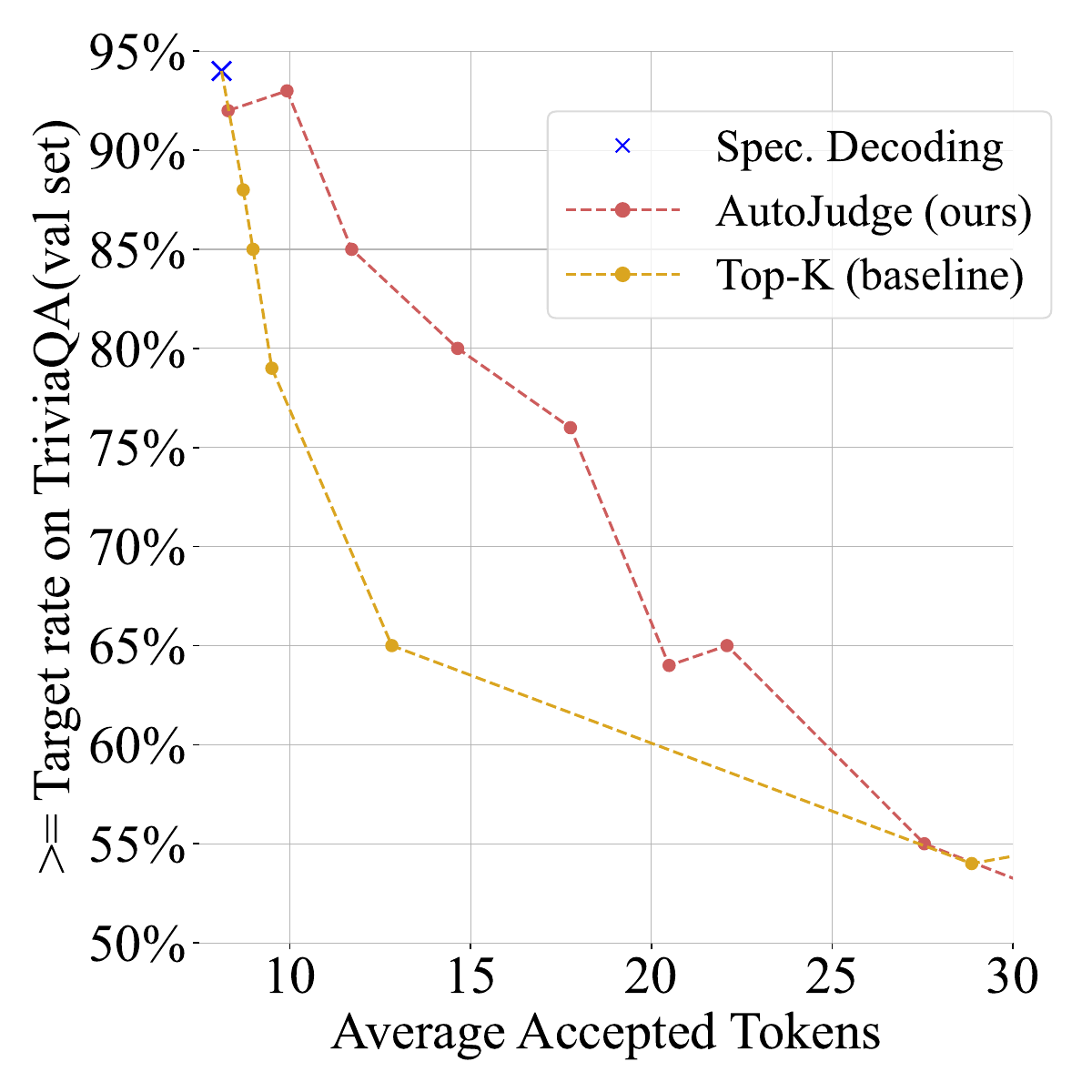} 
    \includegraphics[width=0.49\linewidth]{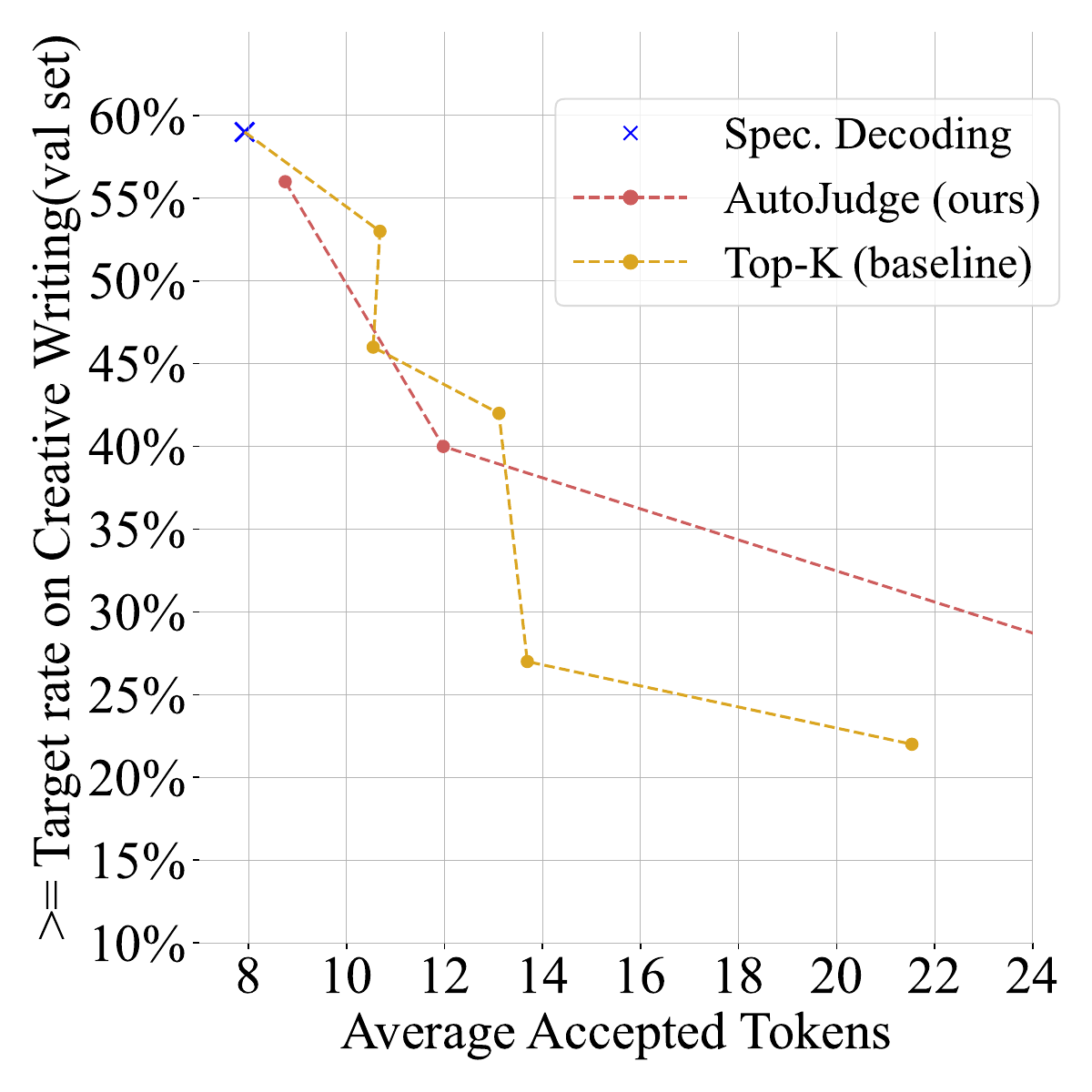} 
    \vspace{-5px}
    \caption{AutoJudge performance on open-ended generation with LLM-as-a-judge evaluation. (Left) closed-book generative question answering on TriviaQA questions (Right) creative writing on Arena-hard-Auto-v2.0 queries. Both use Llama-3.2-1B draft / Llama-3.1-8B target models (Instruct). Creative writing was later found to have noisy labels (>70\% mismatches judged non-equivalent).}
    \label{fig:app_open_ended}\
    \vspace{10px}
\end{figure}

\textbf{Results.}
We report our evaluations on question answering and creative writing in Figure~\ref{fig:app_open_ended} left and right, respectively. Similarly to Sections~\ref{sect:exp_math_gsm8k} and~\ref{sect:exp_coding_lcb}, we evaluate AutoJudge against the top-K baseline in terms of i) the average number of accepted tokens for greedy decoding and ii) the rate at which our outputs were equal or better than the target model. We use the same Claude Sonnet 4 judge and best of 3 voting as specified above. Note that the lossless decoding results are not exactly at 0.5 quality because some response pairs were deemed equal, i.e. [[A=B]].

The results suggest that AutoJudge classifier outperforms traditional speculative decoding on question answering, but not on creative writing. Upon closer inspection, we found that the creative writing benchmark has a very high rate of important tokens --- over 70\% mismatching tokens result in non-equivalent answers, much larger than all other problems. We attribute this anomaly to creative writing being inherently subjective, resulting in noisy LLM judgements. With such high rate of important tokens, there is little room for speed up with AutoJudge. To summarize, we found that AutoJudge can be indeed be used for open-ended problems with the LLM-as-a-judge paradigm, outperforming the baseline in some but not all cases.

\section{Trained Classifier Transfer Between Tasks}\label{app:math_transfer}

\subsection{From GSM8K to MATH-hard Subset}

To evaluate the task transfer capability of AutoJudge, we test whether a classifier trained on one mathematical reasoning dataset can generalize to a different, more challenging dataset. For this evaluation, we use the AutoJudge classifier trained on GSM8K 0-shot with the Llama-3.2-1B-Instruct draft / Llama-3.1-8B-Instruct target model pair (from Section \ref{sect:exp_math_gsm8k}) and apply it to the MATH-hard subset from the LLama Codebook \footnote{We reproduce the setup and get the problems from  \href{https://github.com/meta-llama/llama-cookbook/tree/2501f519c7a775e3fab82ff286916671023ca9c6/tools/benchmarks/llm_eval_harness/meta_eval}{the official Meta evaluations repository}.} based on the lm-evaluation-harness benchmark \citep{eval-harness}.
The MATH-hard dataset contains significantly more challenging mathematical problems compared to GSM8K, requiring more advanced mathematical reasoning. We follow the standard evaluation protocol from the lm-evaluation-harness framework and report exact match accuracy as the downstream metric.

The results are presented in Figure \ref{fig:app_task_transfer_and_long_context} (left). While AutoJudge can still accept more tokens than standard speculative decoding, the accuracy-speed trade-offs are less favorable than the top-K baseline across most operating points. This suggests that the patterns of important tokens learned from grade-school math problems (GSM8K) might not transfer well to more advanced math reasoning tasks.

\subsection{From GSM8K to Long Context}\label{app:long_context}

To test AutoJudge on long-context tasks, we construct a synthetic long-context mathematical reasoning benchmark based on GSM8K. For each test problem, we concatenate 250 randomly sampled GSM8K questions (without their solutions) as context, followed by a single target question that the model must solve. The resulting prompts contain approximately 8-10K tokens before applying the chat template, significantly longer than the standard GSM8K setup.

\begin{figure}[h]
    \centering
    \includegraphics[width=0.49\linewidth]{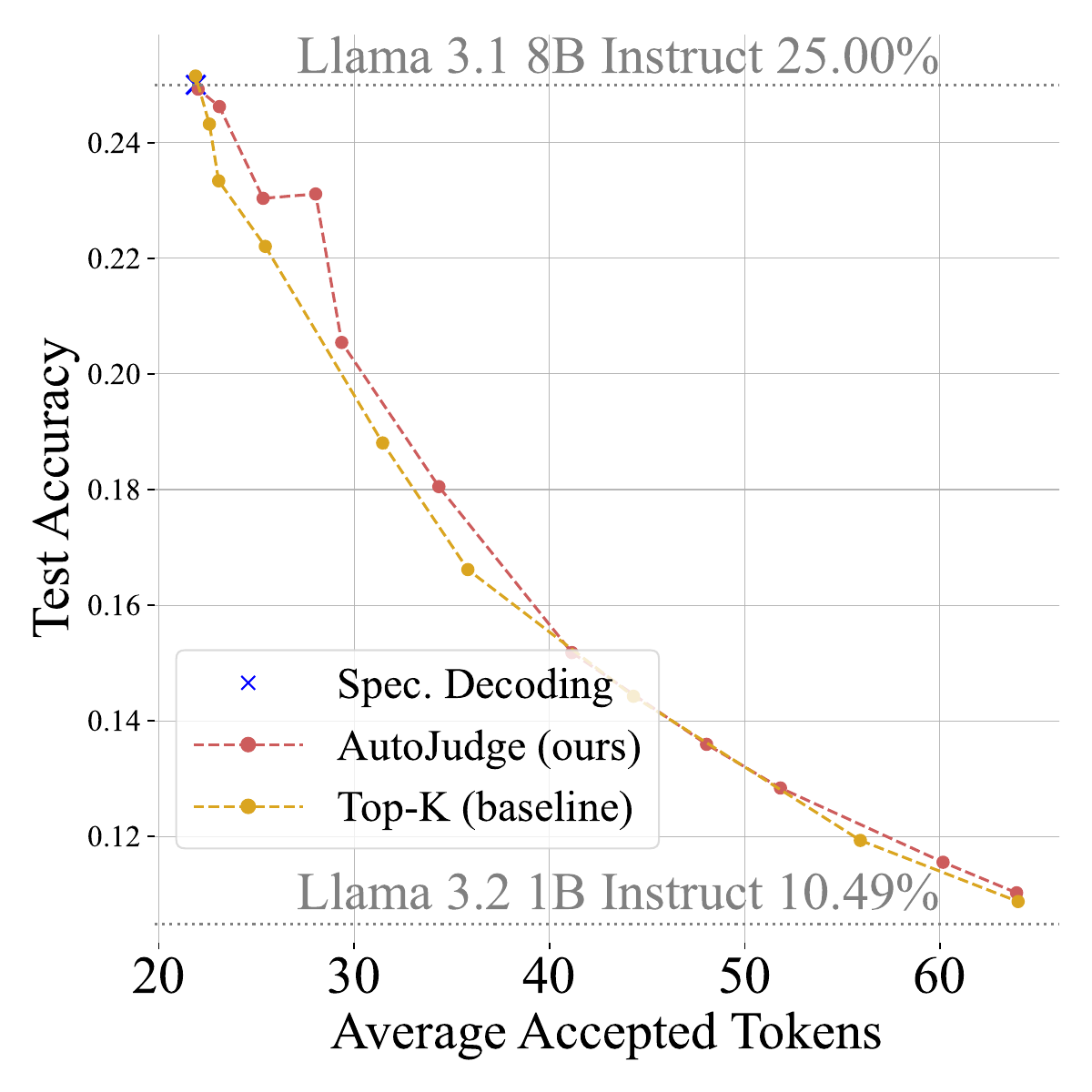} \includegraphics[width=0.49\linewidth]{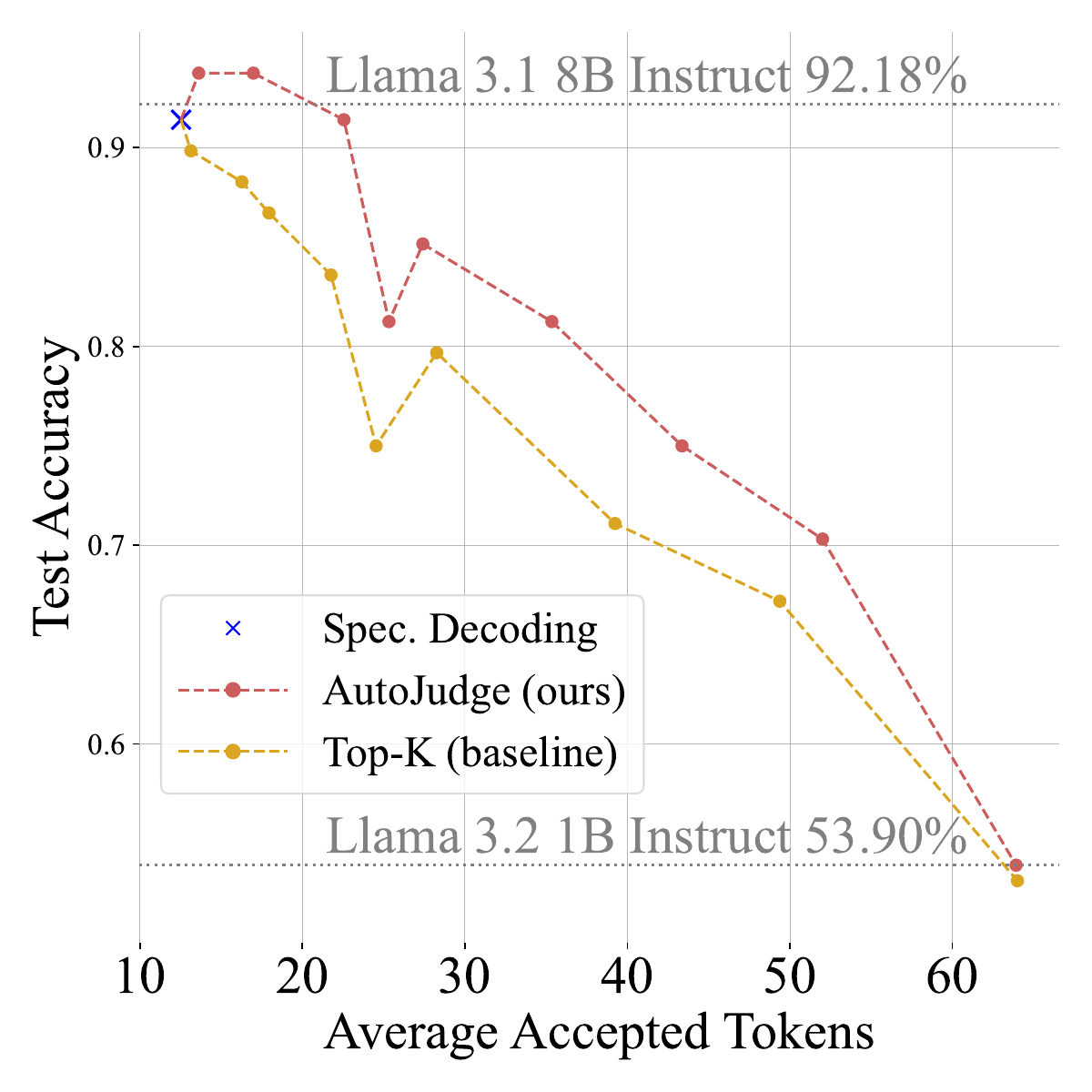}
    \vspace{-5px}
    \caption{(Left) Task transfer: accuracy vs. accepted tokens on MATH-hard using GSM8K-trained classifier. (Right) Long-context performance: accuracy vs. accepted tokens on GSM8K with 8-10K token prompts. Both use Llama-3.2-1B draft / Llama-3.1-8B target setup (Instruct).}
    \label{fig:app_task_transfer_and_long_context}
    \vspace{10px}
\end{figure}

We evaluate the Llama-3.2-1B-Instruct draft / Llama-3.1-8B-Instruct target model pair using the AutoJudge classifier trained on hiddens mined described long context GSM problems. The results are presented in Figure \ref{fig:app_task_transfer_and_long_context} (right). AutoJudge maintains its ability to accept more tokens than standard speculative decoding in this long-context regime, achieving better accuracy while maintaining a similar number of accepted tokens compared to the Top-K baseline. 

These results demonstrate that AutoJudge classifiers trained on standard-length examples can generalize to significantly longer contexts, suggesting that the notion of ``important tokens'' for mathematical reasoning remains consistent regardless of the input length.

\section{Manual Annotation}\label{app:manual_annotation}

\begin{figure}[ht]
    \centering
    \includegraphics[width=0.8\linewidth]{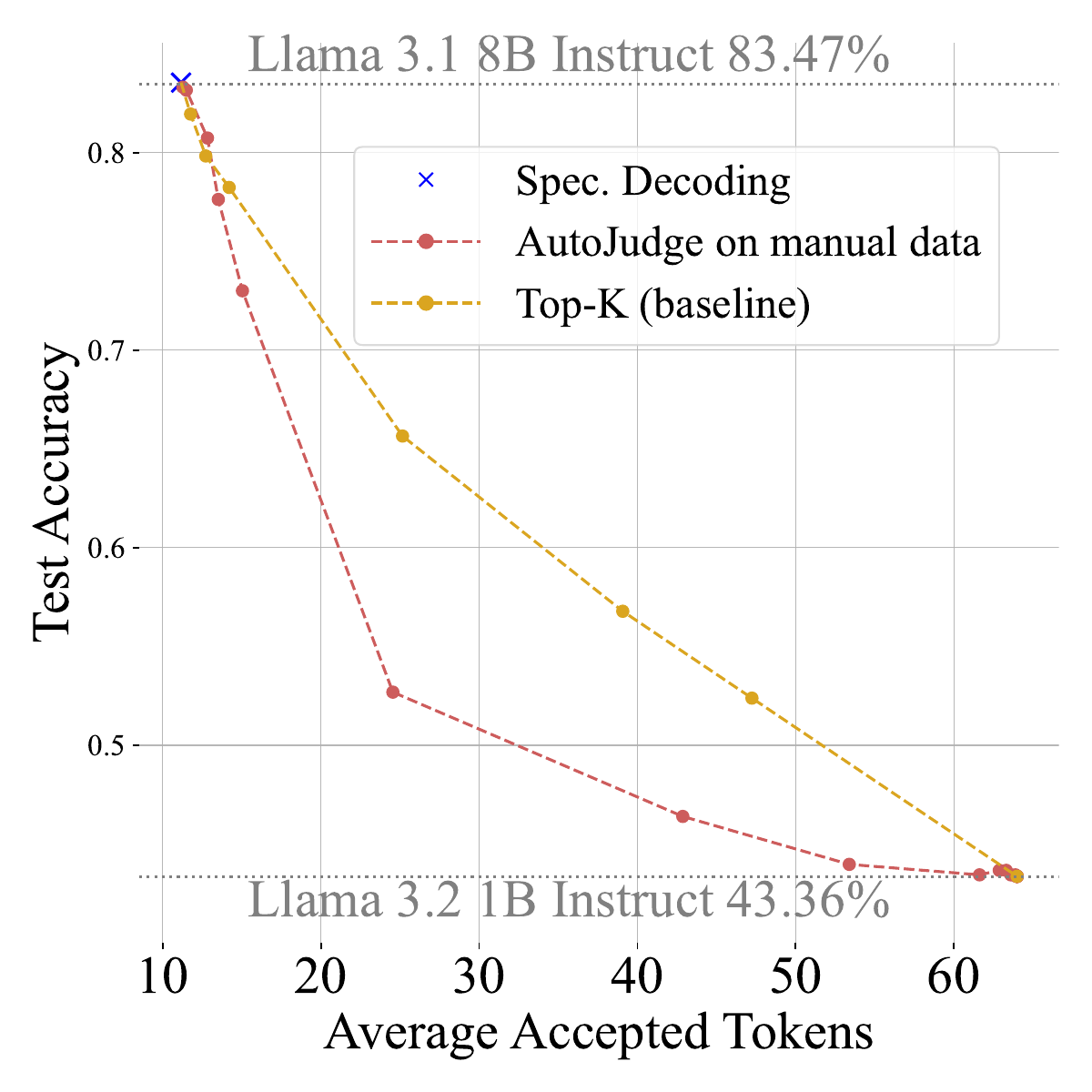}
    \caption{GSM8K 0-shot evaluation of Llama-3.2-1B draft / Llama-3.1-8B target model pair comparing AutoJudge with manual annotation mining (following \citep{bachmann2025judgedecoding}) against Top-K baseline.}
    \label{fig:manual_annotation}
\end{figure}

To compare our automated approach with the manual annotation procedure from Judge Decoding (\cite{bachmann2025judgedecoding}), we conduct an experiment following their methodology as closely as possible. Unfortunately, the original paper does not provide public access to their annotated dataset or source code, which limits direct comparison. Therefore, we recreate their data collection and training procedure based on the methodology described in Section 4.1 of their paper.

\paragraph{Dataset construction.} We curate a dataset of 167 high-quality questions from two sources: the \texttt{ARC-Challenge} benchmark \cite{arc_allenai} (25 questions) and the \texttt{Alpaca} dataset \cite{alpaca} (142 questions, split between 71 mathematical reasoning problems and 71 coding problems). Since the exact question indices used in the original Judge Decoding work are not publicly available, we manually selected challenging and diverse questions from each dataset that represent a range of difficulty levels similar to those described in the original paper.

For each question, we generate responses using three different models from the Llama family: Llama-3.2-1B-Instruct, Llama-3.1-8B-Instruct, and Llama-3.1-70B-Instruct. This produces 501 total generated responses with varying quality levels, providing a diverse set of correct and incorrect continuations for training.

\paragraph{Manual annotation.} Following the procedure described in "Dataset Curation" in Section 4.1 of \citep{bachmann2025judgedecoding}, we manually annotate each generated response to identify the precise location where errors occur. We encode all responses using the Llama-3.1-8B-Instruct model to extract hidden state representations, then manually label each token as either correct or incorrect based on whether accepting that token would lead to a wrong final answer. This annotation process required approximately 8-10 hours of careful manual review by the authors.

\paragraph{Training and evaluation.} Using these manually annotated hidden states, we train AutoJudge classifier with the same architecture and hyperparameters described in Section \ref{sect:exp_math_gsm8k} and Appendix \ref{app:classifier_inputs}. We then evaluate the resulting "manually-annotated" AutoJudge classifier on GSM8K 0-shot with the Llama-3.2-1B-Instruct draft / Llama-3.1-8B-Instruct target setup, comparing against Top-K baseline.
The results are presented in Figure \ref{fig:manual_annotation}. We observe that the manually-annotated classifier performs substantially worse than Top-K baseline and, consecutively, our automatically-mined approach. The manually-annotated classifier achieves lower acceptance rates while exhibiting greater accuracy degradation, suggesting that manual annotation is both less effective and more labor-intensive. We attribute this gap to several factors: (1) human annotators may inconsistently label edge cases or fail to account for how subsequent tokens interact, (2) the manual process is prone to errors over hundreds of annotations, and (3) our automated search procedure (Algorithm \ref{alg:important_tokens_mining}) systematically tests token combinations in a way that better reflects actual inference conditions. These results validate that our automated mining procedure not only eliminates the substantial human effort required for manual annotation (8-10 hours in our case), but also produces more robust training data that leads to better classifier performance.

\section{Correlation between Draft Model and AutoJudge Classifier Probabilities}

One reasonable heuristic for speculative decoding is to consider a token important if a model is certain in its generation, measured in terms of that token's probability. Here, we check whether the AutoJudge classifier is similar to that heuristic. To that end, we measure the correlation between the Llama 3.2 1B Instruct draft model’s probability of the chosen draft token and the corresponding AutoJudge classifier probability on the 1B/8B model pair using the GSM8K dataset (0-shot setting). The resulting Pearson correlation varies within approximately $\pm0.3$ per generation, and the full-sample covariance is $-0.073$. This suggests that AutoJudge classifier does not act on that heuristic.

\section{Hardware Configurations}\label{app:hardware}

We run our experiments primarily on A100-SXM4 GPUs with 80GB DRAM on servers with  dual Epyc 7742 CPU and 1TiB RAM. For 8B/405B model pair we used 8 NVIDIA H100-SXM5 80GB GPUs.

For model pairs that do not fit on a single GPU, we use distributed inference with naive model parallelism (\texttt{device\_map=``auto''}) when using \texttt{transformers} and tensor parallelism for vLLM experiments.

The actual time per experiment varies by the dataset and model pair: 1B draft / 8B target model pair takes, on average, 65.6 seconds to process a GSM8K example, whereas the 8B draft / 70B target takes up an average of 706.4 seconds per sample. On LiveCodeBench, the same 8B draft / 70B target model takes up 449 seconds per sample. Since Algorithm~\ref{alg:important_tokens_mining} can run independently for each sample, we were able to run our code on low-priority preemptible hardware. However, this also makes it hard to measure the exact amount of computations used in our experiments since some of them were lost due to preemption. For running on a single A100/H100 server, please refer to the time per sample above to estimate the total runtime requirements. Please also note that there are ways to mine important tokens more efficiently using APIs (below).

\section{Power consumption}\label{app:energy}
To estimate the energy consumption of AutoJudge, vanilla speculative decoding, and sequential decoding, we run each inference method on a 10\% sample of the GSM8K test set using vLLM. We measure real-world power usage on Llama 3.2 1B draft / 3.1 8B target models with a single A100-SXM4-80GB GPU (Watts) as reported by nvidia-smi, and multiply it by the mean inference time. This represents the GPU-reported power consumption, before adjusting for PSU inefficiency (equally for AutoJudge and baselines) and will vary between GPU types. For convenience, we convert all results to kJ, similar to \cite{maliakel2025investigatingenergyefficiencyperformance}.

\begin{table}[h!]
\centering
\caption{Estimated energy consumption of running inference on 10\% of GSM8K test set.}
\label{tab:energy_consumption}
\vspace{5px}
\begin{tabular}{lccc}
\toprule
\textbf{Method} & \textbf{Autoregressive} & \textbf{Speculative Decoding} & \textbf{AutoJudge} \\
\midrule
\textbf{Power Usage (kJ)} & 87 & 43 & 37 \\
\bottomrule
\end{tabular}
\end{table}

\section{Scaling Up Algorithm~\ref{alg:important_tokens_mining} with API Calls}\label{app:hardware_api}

We would also like to mention that it is possible to scale up the important token discovery in AutoJudge by reframing it in terms of API calls. Note that the search algorithm only ever runs regular generation (greedy or sampling) with the target model and runs parallel forward pass on $\theta_{draft}$.

Hence, we can run Algorithm~\ref{alg:important_tokens_mining} by replacing \texttt{GENERATE(\dots)} on lines 4 and 10 with a call to an LLM generation API with the specified input. With this reframing, we can mine important tokens by querying LLM API providers even if they cannot inference the large target model locally. This can help in a number of use cases: for instance, when developing a speculative decoding algorithm for use with offloading~\citep{specexec,specinfer}.

Additionally, modern open-source inference libraries for LLMs often expose an an OpenAI-compatible API. This way, one can run the important token mining algorithm efficiently over a pool of LLM inference servers, taking advantage of server-side batching and optimized kernels.

Note, however, that this regime requires tokenizing and detokenizing the target model's messages received from API calls, since most public services operate on non-tokenized text strings. Since there are several ways to spell the same text with a given BPE merge table, this can sometimes lead to unintentional ``token healing''.

In our repository, we provide a variant of the important token mining algorithm that leverages the Together Inference API to run the target model. This enables training the AutoJudge classifier even on models that cannot be hosted locally. As a demonstration, we conducted an experiment using Llama 3.1-405B-Instruct (accessed via API) as the target model and Llama 3.1-8B-Instruct as the mining model. The experiment required $\approx$ \$800 in API credits. This example illustrates that Algorithm~\ref{alg:important_tokens_mining} is applicable to models available exclusively through an API. Results obtained for this model pair are presented in Table ~\ref{tab:main_vllm_0shot_8b_405b_and_8b_70b_offloading} (left).

\section{Dataset and Model Licenses}\label{app:extra_formalities_licenses}

In this section, we summarize our use of licensed models and datasets.\begin{itemize}[leftmargin=*]
    \item The GSM8K benchmark~\citep{cobbe2021gsm8k} is licensed under the MIT License;
    \item The LiveCodeBench benchmark~\citep{jain2024livecodebenchholisticcontaminationfree} is licensed under the MIT License;
    \item Llama 3.1 and 3.2 models~\cite{dubey2024llama} are under the Llama Community License Agreement.
\end{itemize}

Our work would also be impossible without open-source software including (but not limited to) PyTorch~\citep{pytorch}, Hugging Face Transformers~\citep{wolf2019huggingface}, vLLM~\citep{kwon2023efficient} and hundreds of other open-source packages in the Python data science \& machine learning ecosystem. Enumerating and acknowledging all these individual packages would take up several pages, but they can be recovered automatically from our repository.

\end{document}